\documentclass[lettersize,journal]{IEEEtran}

\usepackage{amsmath,amsfonts}
\usepackage{array}
\usepackage[caption=false,font=normalsize,labelfont=sf,textfont=sf]{subfig}
\usepackage{textcomp}
\usepackage{stfloats}
\usepackage{xurl}
\usepackage{graphicx}
\usepackage{cite}
\usepackage{booktabs}
\usepackage{forest}
\usepackage{tikz}
\usetikzlibrary{arrows.meta}

\begin{document}

\title{GeoFormer: A Lightweight Swin Transformer for Joint Building Height and Footprint Estimation from Sentinel Imagery}

\author{Jinzhen~Han,
        JinByeong~Lee,
        Jisung~Kim,
        MinKyung~Cho,
        DaHee~Kim,
        and~HongSik~Yun,~\IEEEmembership{Member,~IEEE}
\thanks{This research was supported by the 2023-MOIS36-004 (RS-2023-00248092) of the Technology Development Program on Disaster Restoration Capacity Building and Strengthening funded by the Ministry of Interior and Safety (MOIS, Korea).}%
\thanks{J.~Han, J.~Lee, and H.~Yun are with the Department of Civil, Architectural and Environmental System Engineering, Sungkyunkwan University, South Korea (e-mail: yoonhs@skku.edu).}%
\thanks{J.~Kim is with the School of Geography, University of Leeds, United Kingdom.}%
\thanks{M.~Cho and D.~Kim are with the Interdisciplinary Program for Crisis, Disaster and Risk Management, Sungkyunkwan University, South Korea.}}

\markboth{IEEE Journal of Selected Topics in Applied Earth Observations and Remote Sensing}%
{Han \MakeLowercase{\textit{et al.}}: GeoFormer for Building Height and Footprint Estimation}

\maketitle

\begin{abstract}
Building height (BH) and footprint (BF) are fundamental urban morphological parameters required by climate modelling, disaster-risk assessment, and population mapping, yet globally consistent data remain scarce.
In this work, we develop \textbf{GeoFormer}, a lightweight Swin Transformer-based multi-task learning framework that jointly estimates BH and BF on a 100\,m grid using only open-access Sentinel-1 SAR, Sentinel-2 multispectral, and DEM data.
A geo-blocked data-splitting strategy enforces strict spatial independence between training and evaluation regions across 54 morphologically diverse cities.
We set representative CNN baselines (ResNet, UNet, SENet) as benchmarks and thoroughly evaluate GeoFormer's prediction accuracy, computational efficiency, and spatial transferability.
Results show that GeoFormer achieves a BH RMSE of 3.19\,m with only 0.32\,M parameters---outperforming the best CNN baseline (UNet) by 7.5\,\%---indicating that windowed local attention is more effective than convolution for scene-level building-parameter retrieval.
Systematic ablation on context window size, model capacity, and input modality further reveals that a 5$\times$5 (500\,m) receptive field is optimal, DEM is indispensable for height estimation, and multispectral reflectance carries the dominant predictive signal.
Cross-continent transfer tests confirm BH RMSE below 3.5\,m without region-specific fine-tuning.
All code, model weights, and the resulting global product are publicly released.
\end{abstract}


\section{Introduction}

\IEEEPARstart{I}{n} recent decades, urban areas across the globe---especially in Asia---have undergone dramatic three-dimensional transformations driven by population growth, land scarcity, and vertical densification. Studies have documented not only horizontal expansion but also a shift toward vertical urban development, particularly in megacities such as Beijing and Seoul~\cite{fang2016changing, frolking2013global, frolking2024global}. This transition has profound implications for urban climate, infrastructure, and risk management.

Building height (BH) and building footprint (BF) are critical indicators of urban form. Accurate BH and BF data are essential for modelling urban heat islands~\cite{xi2021impacts, perini2014effects}, estimating flood exposure~\cite{huang2020estimates}, and simulating cascading disasters such as fire-following-earthquakes~\cite{tian2025fire}. However, despite their significance, public access to consistent and high-resolution 3D urban data remains limited---particularly in the Global South~\cite{geofabrik_osm_2018}. Even in data-rich regions such as South Korea, recent surveys show that only 30--40\% of buildings include height attributes.

Numerous methods have been developed to estimate BH and BF, spanning rule-based and empirical regression models~\cite{li2020developing, cai2023deep}, classical machine learning~\cite{frantz2021national, dabrock2024leveraging}, and deep learning architectures~\cite{buyukdemircioglu2022deep, li2021deep, shafts2023}. These approaches exhibit a distinct trade-off between accuracy, data dependency, and spatial scalability. The highest per-building accuracies have been achieved using very-high-resolution (VHR) optical imagery~\cite{buyukdemircioglu2022deep, li2021deep, rastogi2022automatic}, airborne LiDAR~\cite{park2019creating, li2020developing}, or VHR SAR~\cite{sun2019large}. While these methods can reach RMSE values well below 2\,m for individual buildings, they are inherently constrained by the high cost, limited spatial coverage, and infrequent update cycles of the underlying data, making consistent global-scale deployment impractical~\cite{cai2024automated}.

Open-access Sentinel-1 SAR and Sentinel-2 MSI imagery have enabled grid-based estimation at broader scales~\cite{frantz2021national, wu2023first, cai2023deep, chen2024refining}. However, many of these models are trained and evaluated over a single city or a small number of regions, leaving their transferability to morphologically different urban environments largely unverified. Moreover, several rely on auxiliary layers---cadastral footprints, land-use maps, or slope data---during inference~\cite{wu2023first, chen2025structure}, which restricts applicability wherever such ancillary information is unavailable or inconsistent. Traditional ML pipelines further require manual feature engineering~\cite{dabrock2024leveraging, che20243d}, limiting their capacity to capture complex spatial--spectral relationships and reducing adaptability across sensor configurations.

A new wave of studies has sought to push spatial coverage further through advanced architectures and multi-source fusion. \cite{wang2024mfbhnet} introduced MF-BHNet, a hybrid Transformer--CNN encoder that fuses Sentinel-1 and Sentinel-2 at 10\,m resolution, though validation remains limited to a few European sites. \cite{zheng2025nectnet} proposed NeCT-Net, integrating spaceborne LiDAR with Sentinel data via a CNN--Transformer hybrid, yet its dependence on ICESat-2 point clouds limits spatial continuity. \cite{mostafavi2024utglobus} released UT-GLOBUS with per-building heights for over 1\,200 cities, but the method requires pre-existing building footprint inventories (e.g., OpenStreetMap) as input---a resource that remains incomplete in much of the Global South. At the finest scale, \cite{zhu2025globalbuildingatlas} mapped 2.75 billion buildings at 3\,m resolution; however, this product depends on commercial PlanetScope imagery rather than freely available sensors. These advances---while impressive---underscore a persistent trade-off: higher spatial resolution or broader coverage invariably comes at the expense of additional proprietary data, specialised reference layers, or restricted geographic validation. Across all existing work, most Sentinel-based models address either BH or BF in isolation, without exploiting the well-documented physical coupling between the two variables, and few have been systematically validated across multiple countries or markedly different urban morphologies.

The choice of 100\,m grid resolution in this work is motivated by three mutually reinforcing arguments. First, at 10\,m resolution a single Sentinel pixel frequently straddles multiple buildings, inter-building gaps, cast shadows, and vegetation~\cite{fisher1997pixel, weng2012remote}. In SAR imagery, this mixing is further compounded by double-bounce scattering and layover from adjacent structures, effects that are well documented even at very-high resolution and persist at the 10\,m Sentinel-1 scale~\cite{brunner2010earthquake}. \cite{schug2022sub} showed that even advanced regression-based unmixing of Sentinel-1/2 data systematically underestimates built-up areas in densely packed urban cores, while \cite{radoux2016sentinel} reported that landscape features smaller than the effective spatial sampling unit of Sentinel-2 cause persistent sub-pixel contamination. This is particularly severe in high-density Asian and European cities where inter-building separations are $\le$10\,m. Aggregation to a 100\,m ``scene'' integrates over these sub-pixel artefacts, yielding a physically more meaningful and statistically more robust measure of mean building height and fractional footprint coverage. Second, as summarised in Table~\ref{tab:global_products}, every operational global building-height or built-up product published to date adopts a grid resolution of $\sim$90--250\,m. This convergence is not coincidental: at 100\,m, reliable building-parameter retrieval can be achieved using only freely available Sentinel and DEM data, eliminating the need for proprietary VHR imagery or ancillary vector layers. Operating at this resolution therefore allows GeoFormer to maintain full global reproducibility while remaining directly comparable to the established product ecosystem.

\begin{table}[!t]
\centering
\small
\setlength{\tabcolsep}{2pt}
\caption{Spatial resolution of operational global building-height and built-up products.}
\label{tab:global_products}
\begin{tabular}{lccp{2.5cm}}
\toprule
\textbf{Product} & \textbf{Resolution} & \textbf{Year} & \textbf{Source} \\
\midrule
WSF3D (DLR)           & 90\,m   & 2022 & TanDEM-X + S1/S2 \\
GHS-BUILT-H (JRC)     & 100\,m  & 2023 & AW3D30 + SRTM + S2 \\
SHAFTS                & 100\,m  & 2023 & Sentinel + vector data \\
GBH2020               & 150\,m  & 2024 & GEDI + S1/S2 + Landsat \\
GHS-POP (JRC)         & 100\,m / 1\,km & 2019 & GHS-BUILT input \\
GHSL-SMOD (JRC)       & 1\,km   & 2023 & GHS-BUILT input \\
\textbf{GeoFormer (ours)} & \textbf{100\,m} & \textbf{2025} & \textbf{S1 + S2 + DEM} \\
\bottomrule
\end{tabular}
\end{table}

Third, the 100\,m grid directly serves the major downstream user communities: the WUDAPT/Local Climate Zone framework operates at 100\,m~\cite{stewart2012local, demuzere2022global, ching2018wudapt, bechtel2015mapping}, mesoscale weather models ingest urban canopy parameters at $\ge$100\,m~\cite{martilli2002urban}, global population grids such as WorldPop and GHS-POP are disseminated at 100\,m~\cite{tatem2017worldpop, schiavina2023ghspop}, and urban building energy models aggregate at the district/block level ($\sim$100--250\,m)~\cite{reinhart2016urban}. A finer-resolution product would need to be re-aggregated before use in any of these pipelines, adding processing overhead without improving the downstream result. One might further ask whether a 10\,m model could simply be aggregated to 100\,m; however, pixel-level loss functions do not optimise scene-level aggregation error, so biases can amplify upon spatial averaging; global 10\,m processing incurs $\sim$$10^{4}\times$ the computational cost of 100\,m, prohibiting routine updates for most institutions; and a native 100\,m model can adopt a spatial-neighbourhood context window (e.g., GeoFormer's $5\times5 = 500$\,m) that captures urban-block morphology directly, whereas equivalent context at 10\,m would require $50\times50$-pixel patches per prediction---a configuration that is neither practical nor commonly attempted.

\begin{figure*}[!t]
\centering
\includegraphics[width=0.85\textwidth]{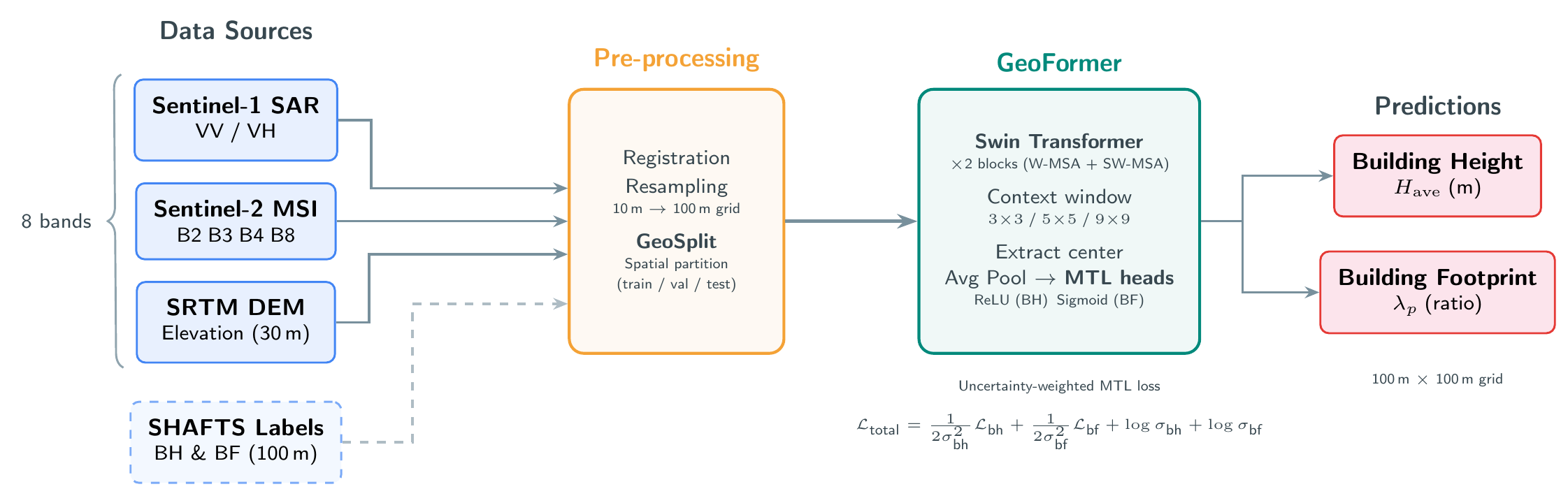}
\caption{Workflow of the proposed GeoFormer framework.}
\label{fig:workflow}
\end{figure*}

To overcome the above limitations, we develop \textbf{GeoFormer}, a lightweight Swin Transformer-based multi-task learning framework for joint building height (BH) and footprint (BF) estimation at 100\,m resolution. GeoFormer relies solely on globally accessible Sentinel-1 SAR, Sentinel-2 MSI, and digital elevation model (DEM) data, avoiding any proprietary VHR imagery or vector sources. As shown in Fig.~\ref{fig:workflow}, BH/BF reference labels are spatially aligned with satellite and DEM inputs on a unified 100\,m grid, and a geo-blocked spatial partitioning strategy (GeoSplit) ensures strict independence between training and test regions. The Swin Transformer backbone, with configurable context windows (3$\times$3, 5$\times$5, 9$\times$9), enables joint BH and BF estimation via multi-task learning.

We set representative CNN architectures (ResNet, UNet, SENet) as benchmarks and thoroughly evaluate GeoFormer in terms of prediction accuracy, computational efficiency, and spatial transferability across 54 morphologically diverse cities. The main contributions of this work are:
\begin{enumerate}
    \item We develop GeoFormer, a compact Swin-based multi-task framework (0.32\,M parameters) that jointly predicts BH and BF from open-source satellite data at 100\,m resolution, achieving a BH RMSE of 3.19\,m---outperforming the best CNN baseline (UNet) by 7.5\,\%.
    \item Through systematic comparison with CNN baselines under identical training conditions, we demonstrate that windowed local attention is more effective than convolution for scene-level building-parameter retrieval, while requiring 35$\times$ fewer parameters than ResNet-18.
    \item Through ablation on context window size, model capacity, and input modality, we establish that a 5$\times$5 (500\,m) receptive field is optimal, DEM is indispensable for height estimation, and multispectral reflectance carries the dominant predictive signal.
    \item All code, model weights, and the resulting global 100\,m BH/BF product are publicly released, enabling reproducibility and downstream applications in urban climate modelling, disaster-risk assessment, and population mapping.
\end{enumerate}

\section{Data Preprocessing}

\subsection{Reference Data}

This study adopts the open-source SHAFTS Reference Dataset (v2022.3)~\cite{shafts2023}, which provides per-city building footprint and height metrics that have already been processed using Fishnet Analysis~\cite{musiaka2021}. Specifically, a regular 100\,m~$\times$~100\,m grid was applied to vector building inventories by the SHAFTS authors to compute standardized spatial indices. The underlying vector building inventories originate from authoritative cadastral surveys and open-source datasets (e.g., OpenStreetMap), and the derived grid-level labels were validated against LiDAR-based DSM references in representative cities~\cite{shafts2023}. These precomputed grid-based metrics are directly used in this study as reference labels.

\begin{figure}[!t]
  \centering
  \includegraphics[width=0.8\linewidth]{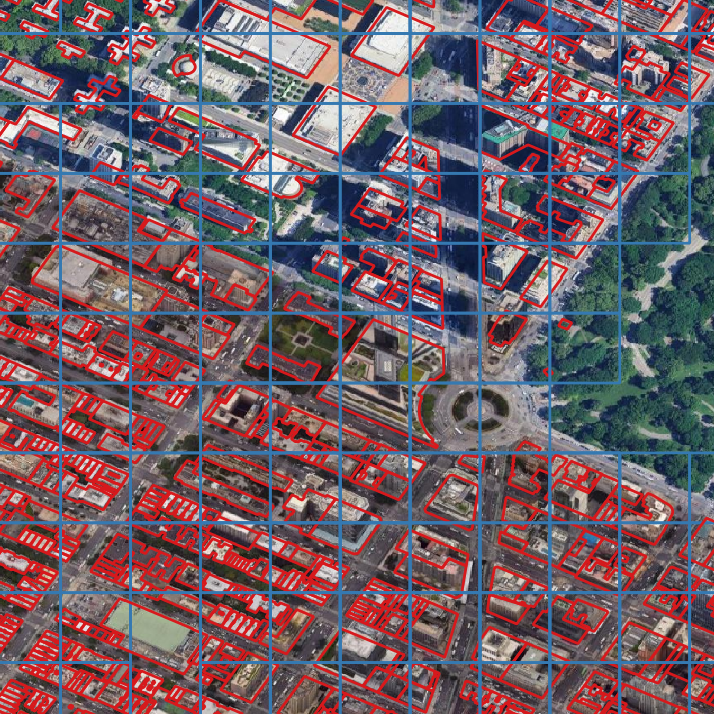}
  \caption{Illustration of Fishnet Analysis: a 100\,m grid overlays vector building footprints to compute per-cell height and footprint coverage.}
  \label{fig:fishnet}
\end{figure}

As shown in Fig.~\ref{fig:fishnet}, Fishnet Analysis discretizes the urban space by overlaying a uniform grid on building polygons. Within each cell, the intersected building areas and their corresponding heights are used to derive cell-level statistics. This allows consistent aggregation of spatial information and enables raster-format learning targets.

The two core indices, building footprint ratio and average building height, are defined as follows:

\begin{align}
\lambda_p &= \frac{\sum_{i\in \mathcal{J}} A_i}{(100\,\mathrm{m})^2}, \\
H_{\mathrm{ave}} &= \frac{\sum_{i\in \mathcal{J}} A_i\,h_i}{\sum_{i\in \mathcal{J}} A_i},
\end{align}

where $\mathcal{J}$ denotes the set of buildings that spatially intersect a given 100\,m grid cell, $A_i$ is the intersection area between the $i$th building and the cell, and $h_i$ is the height of the $i$th building. The footprint index $\lambda_p$ measures the proportion of the cell covered by building structures, while the average building height $H_{\mathrm{ave}}$ is computed as the area-weighted mean of building heights within the cell. Note that both metrics are normalized within the fixed 100\,m~$\times$~100\,m spatial unit, enabling consistent comparisons across diverse urban morphologies.

\begin{figure*}[!t]
  \centering
  \includegraphics[width=\linewidth]{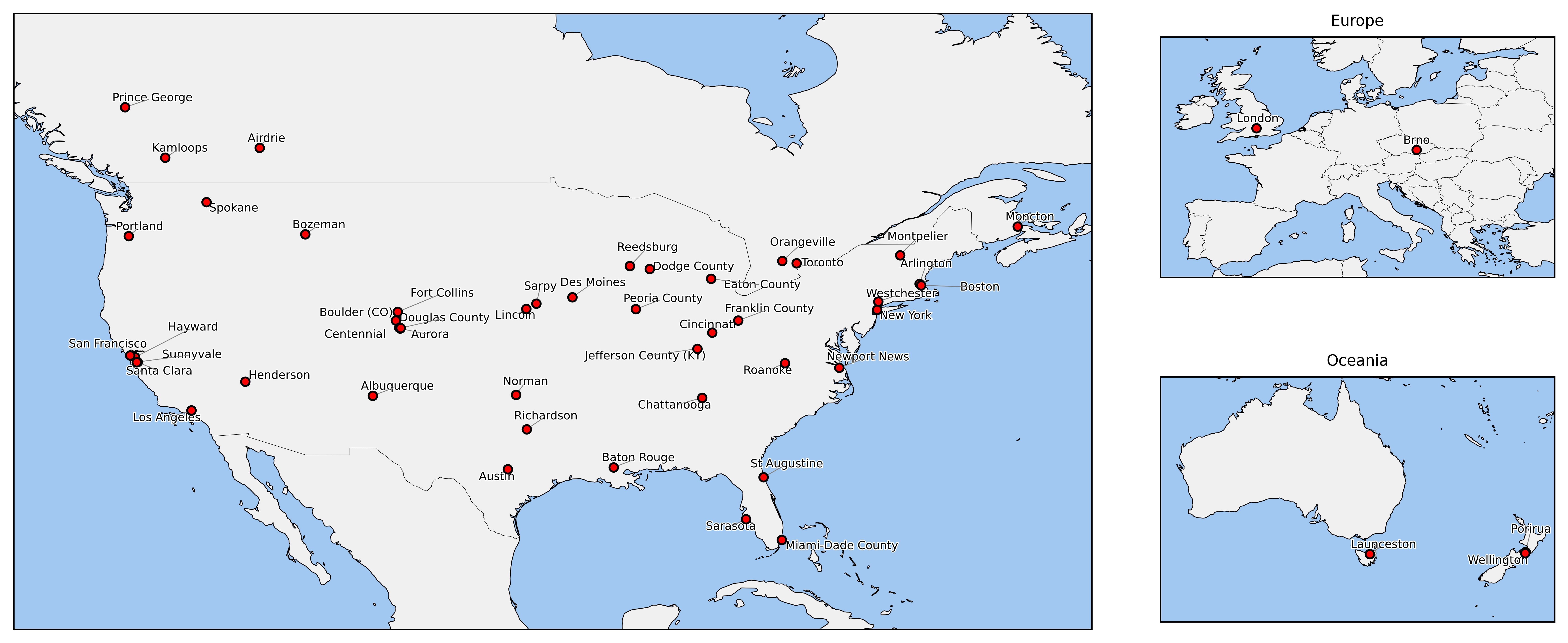}
  \caption{Geographic distribution of SHAFTS (v2022.3) reference cities.}
  \label{fig:shafts_cities}
\end{figure*}

As shown in Fig.~\ref{fig:shafts_cities}, this study uses a total of 51 cities from the SHAFTS dataset that explicitly provide building height attributes. These cities are geographically diverse, covering multiple continents and urban morphologies, and thus offer a broad representation of global built environments for robust model training and evaluation.

\subsection{Explanatory Data and Processing}

This study employs three open geospatial sources as explanatory variables: Sentinel-1 Synthetic Aperture Radar (SAR), Sentinel-2 multispectral optical imagery, and Shuttle Radar Topography Mission (SRTM) digital elevation data. The corresponding specifications are summarized in Table~\ref{tab:data_sources}. Sentinel imagery was processed at 10\,m resolution, while SRTM DEM was used at 30\,m resolution.

\begin{table*}[!t]
  \centering
  \caption{Explanatory Data Sources and Specifications}
  \label{tab:data_sources}
  \begin{tabular}{p{0.3\textwidth} p{0.6\textwidth}}
    \toprule
    \textbf{Data Source} & \textbf{Specification} \\
    \midrule
    Sentinel-1 GRD Level-1 & 10 m VV and VH polarizations \\
    Sentinel-2 Level-2A     & 10 m red, green, blue, and near-infrared (B2, B3, B4, B8) \\
    SRTM V3 DEM             & 30 m digital elevation model \\
    \bottomrule
  \end{tabular}
\end{table*}

SAR backscatter captures surface characteristics via three scattering mechanisms: surface scattering (weak and specular), volume scattering (diffuse return from volumetric targets), and double-bounce scattering (strong signal from vertical structures). Optical reflectance reveals material-specific spectral signatures and surface textures, useful for distinguishing man-made structures. SRTM DEM provides global elevation priors essential for terrain-aware modeling.

To acquire these datasets, we implemented an automated download pipeline on the Google Earth Engine (GEE) platform using JavaScript. For each city, the raster filename from the SHAFTS dataset was parsed to extract its geographic bounding box and acquisition year. These parameters were then used to retrieve Sentinel-1, Sentinel-2, and SRTM imagery. During the download process, additional preprocessing steps were applied, including cloud filtering (for Sentinel-2), annual percentile aggregation, and clipping to the padded city extent. This ensured that the explanatory data were temporally consistent, cloud-free, and spatially aligned across all cities.

To construct a unified dataset suitable for model training and testing, the reference variables (i.e., building height and footprint) were combined with the corresponding explanatory variables (i.e., Sentinel and DEM patches) on a per-sample basis. Prior to this integration, a set of filtering rules was applied to remove implausible or uninformative samples.

Each \(10 \times 10\) pixel patch corresponds to a 100\,m grid cell. Implausible samples were removed based on three rules:
\begin{enumerate}
  \item \(2.0 \le H_{\mathrm{ave}} \le 500.0\) (m),
  \item \(\lambda_p > 0.01\),
  \item If \(\lambda_p < 0.04\), then require \(H_{\mathrm{ave}} < 20\) to exclude slivers.
\end{enumerate}
These criteria ensure that (i) height values lie within a physically plausible range, (ii) the minimum detectable footprint reflects one-pixel coverage, and (iii) narrow artifacts with unrealistically high structures are excluded.

\begin{figure}[!t]
  \centering
  \resizebox{0.9\linewidth}{!}{
    \begin{forest}
      for tree={
        grow=east,
        draw,
        rounded corners,
        node options={align=center},
        edge={-{Latex}},
        parent anchor=east,
        child anchor=west,
        l=1.5cm
      }
      [dataset.h5
        [NewYork
          [reference
            [building\_height\\(100 m grid)]
            [building\_footprint\\(100 m grid)]
          ]
          [explanatory
            [SRTM\\(30 m DEM)]
            [Sentinel-1
              [VV]
              [VH]
            ]
            [Sentinel-2
              [B2]
              [B3]
              [B4]
              [B8]
            ]
          ]
          [coordinates
            [row\_index]
            [col\_index]
          ]
        ]
      ]
    \end{forest}
  }
  \caption{Structure of a single city group in the final HDF5 file.}
  \label{fig:h5_structure}
\end{figure}
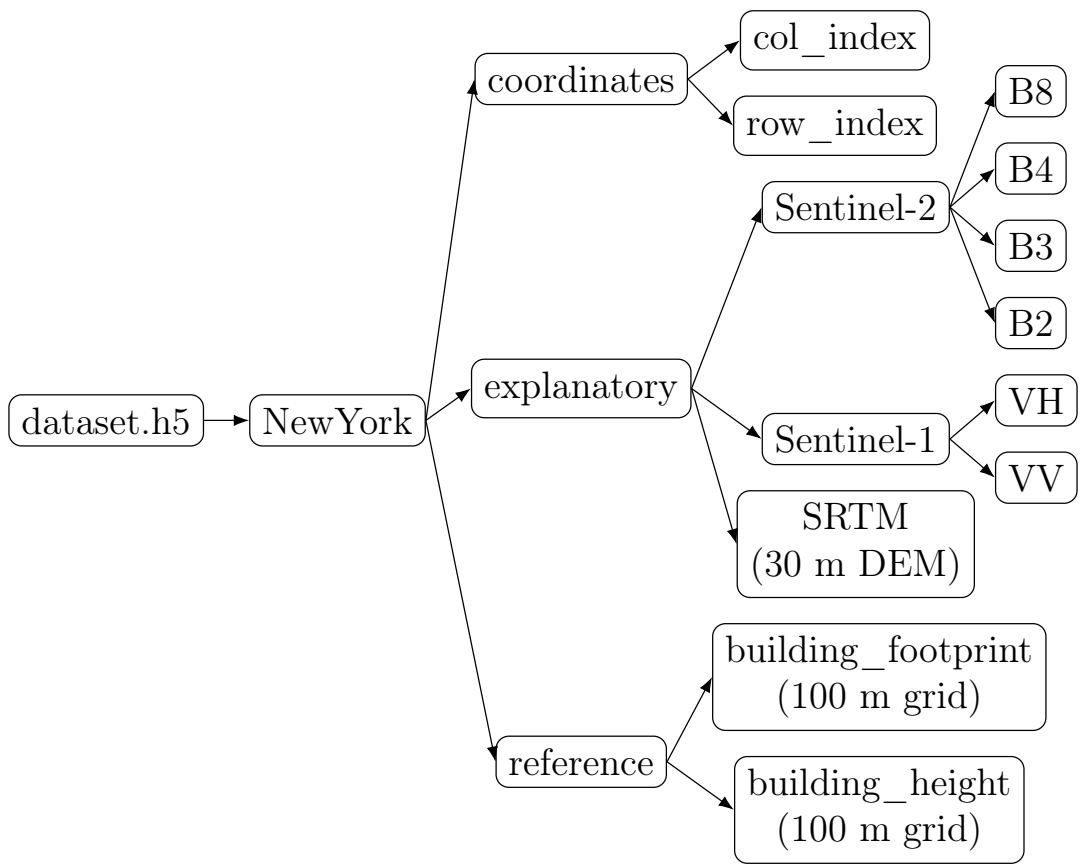

Filtered samples were consolidated into a unified HDF5 file, with each city stored as an individual group. Fig.~\ref{fig:h5_structure} illustrates the internal structure of one such group. Each group contains two types of reference labels (building height and footprint), multi-source explanatory variables (Sentinel-1, Sentinel-2, and DEM), and a pair of coordinate indices.

The added \texttt{row\_index} and \texttt{col\_index} fields represent the grid-based location of each patch within the city, expressed in row--column units. These coordinates were appended after filtering to facilitate spatial operations such as neighborhood-aware patch concatenation during training.

In this study, the dataset is split into training, validation, and test sets using an 8:1:1 ratio. Unlike prior work that relies on fully random sampling, we address key limitations of such approaches, especially in the context of receptive-field-based modeling.

\begin{figure}[!t]
  \centering
  \includegraphics[width=\linewidth]{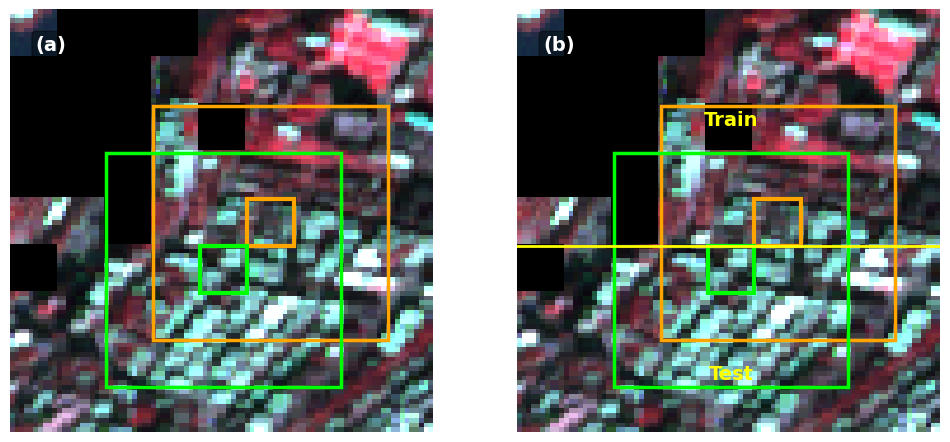}
  \caption{Data leakage from random sampling under dynamic receptive field concatenation.}
  \label{fig:shift}
\end{figure}

As shown in Fig.~\ref{fig:shift}, random partitioning can lead to overlap between training and test patches when dynamic context concatenation is applied at runtime. This overlap may cause the model to unintentionally observe parts of the test region during training, leading to performance inflation.

\begin{figure}[!t]
  \centering
  \includegraphics[width=\linewidth]{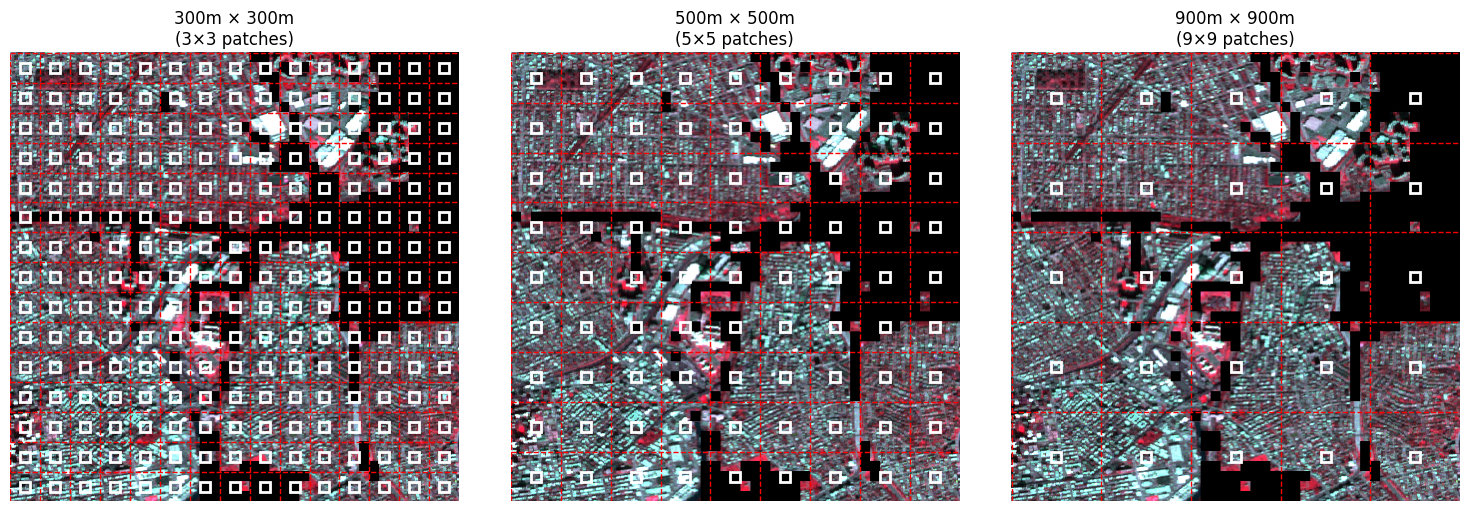}
  \caption{Sample reduction under static receptive field expansion.}
  \label{fig:patches}
\end{figure}

Precomputing context patches in advance can eliminate such leakage. However, as shown in Fig.~\ref{fig:patches}, increasing the receptive field (e.g., from 100\,m to 900\,m) drastically reduces the number of usable samples, making fair cross-scale comparison difficult.

To overcome these issues, we adopt a spatially aware partitioning strategy based on the geographic layout of each city. Instead of random selection, each city is divided into ten equal-angle radial sectors centered on the urban core. These sectors are then allocated to the training, validation, and test sets using a greedy balancing strategy to approximate the desired 8:1:1 split while ensuring each subset contains both central and peripheral regions.

\begin{figure}[!t]
  \centering
  \includegraphics[width=0.8\linewidth]{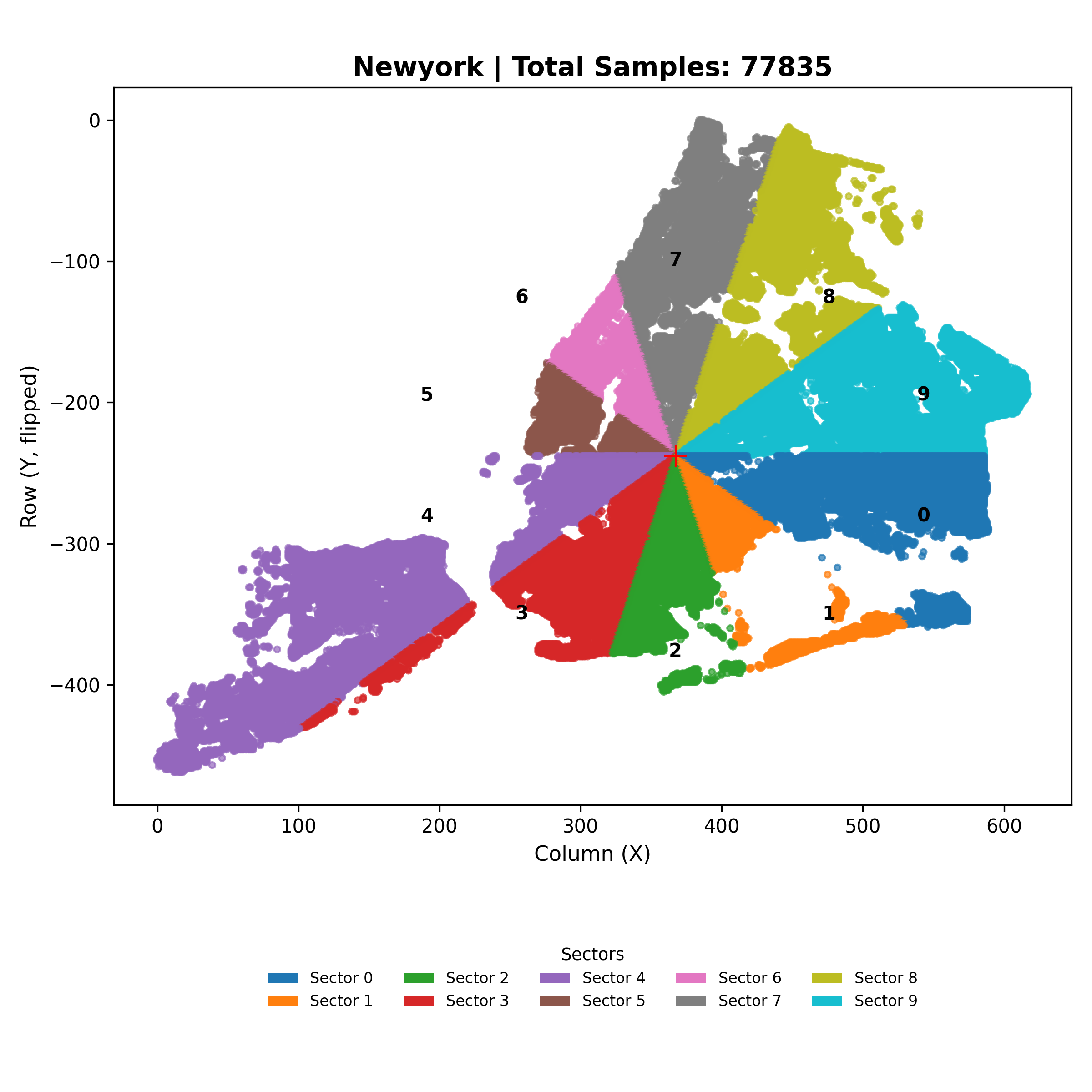}
  \caption{Radial sector division of New York City used for spatially balanced dataset partitioning.}
  \label{fig:nyc_sectors}
\end{figure}

Fig.~\ref{fig:nyc_sectors} illustrates the radial-sector partitioning for New York City. This strategy preserves spatial coherence within each subset and avoids training-test overlap along receptive field boundaries.

\begin{figure}[!t]
  \centering
  \includegraphics[width=\linewidth]{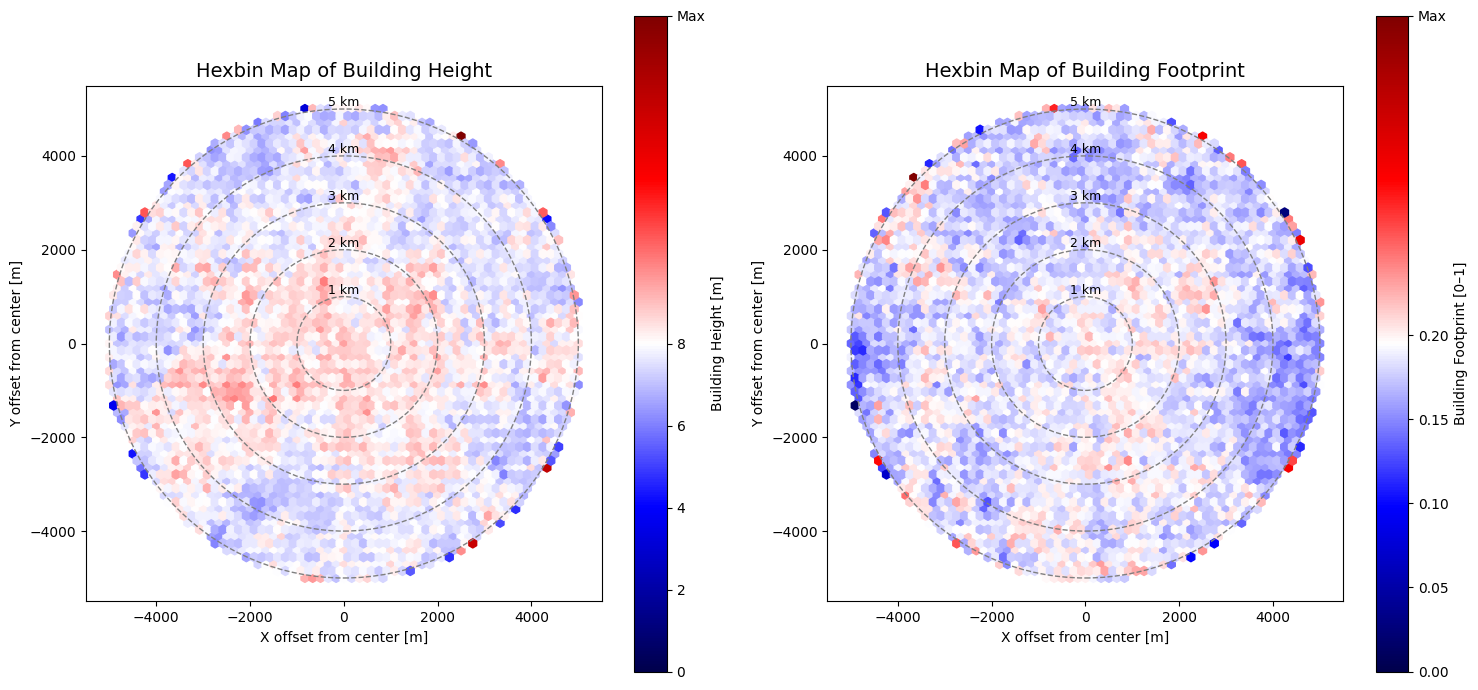}
  \caption{Hexbin-mapped distribution of building height and footprint across training cities.}
  \label{fig:hexbinmap}
\end{figure}

To verify the necessity of spatial partitioning, Fig.~\ref{fig:hexbinmap} presents hexbin visualizations of building height (BH) and footprint (BF) relative to city centers. Both metrics show a clear decline from core to periphery, confirming the spatial heterogeneity of urban form.

\begin{table}[!t]
\centering
\small
\setlength{\tabcolsep}{3pt}
\caption{Ring-based statistics of building height and footprint}
\label{tab:ring_stats}
\begin{tabular}{lcccc}
\toprule
\textbf{Range (m)} & \textbf{H Mean} & \textbf{H Std.} & \textbf{BF Mean} & \textbf{BF Std.} \\
\midrule
0--1000    & 10.13 & 11.03 & 0.203 & 0.132 \\
1000--2000 & 9.58  & 7.85  & 0.200 & 0.130 \\
2000--3000 & 9.05  & 6.75  & 0.195 & 0.122 \\
3000--4000 & 8.64  & 6.65  & 0.190 & 0.124 \\
4000--5000 & 8.43  & 6.68  & 0.188 & 0.126 \\
\bottomrule
\end{tabular}
\end{table}

Table~\ref{tab:ring_stats} further quantifies these gradients using concentric 1\,km bands, showing consistent decreases in BH and BF with radial distance. This validates the need for spatially stratified sampling to maintain generalization under urban heterogeneity.

\section{Methodology}

\subsection{Model Architecture}

\begin{figure*}[!t]
\centering
\includegraphics[width=0.9\textwidth]{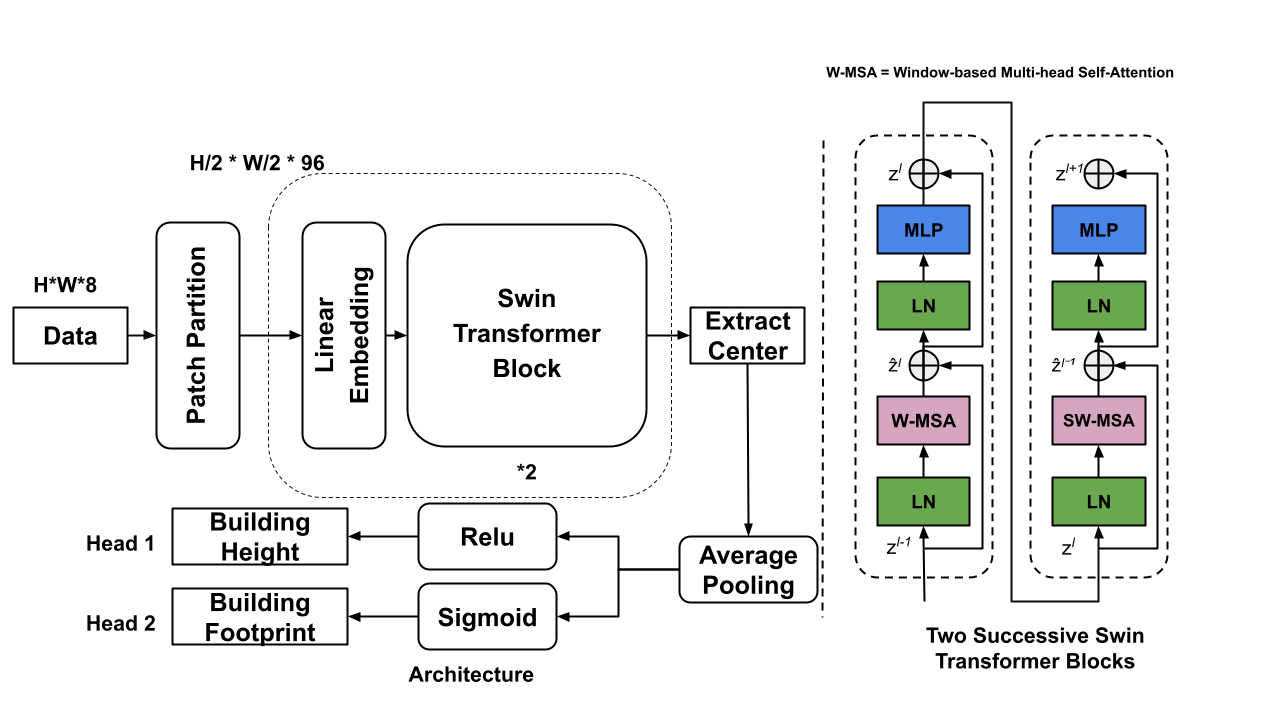}
\caption{Architecture of GeoFormer: a Swin-based multi-task model for predicting building height and footprint.}
\label{fig:model}
\end{figure*}

To estimate both building height (BH) and building footprint (BF) simultaneously, we design a Swin Transformer--based multi-task regression framework named \textbf{GeoFormer}. The complete architecture is depicted in Fig.~\ref{fig:model}, which shows the flow of information from input preparation to final predictions.

\begin{figure*}[!t]
\centering
\includegraphics[width=0.9\textwidth]{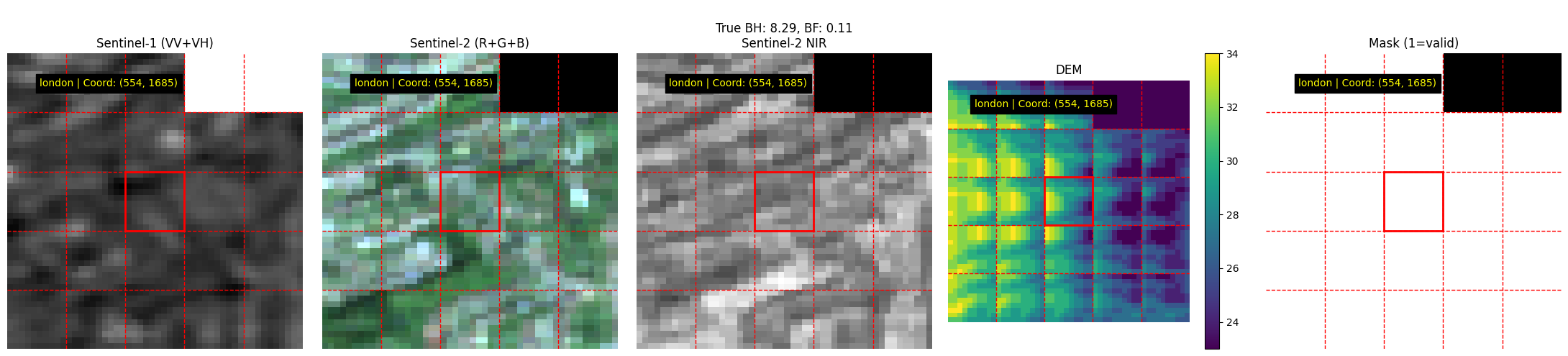}
  \caption{Illustration of the 8-band multi-source input tensor. From left to right: Sentinel-1 (VV, VH), Sentinel-2 (RGB+NIR), true BH, DEM, and the binary mask indicating the valid region (center patch).}
  \label{fig:input_tensor}
\end{figure*}

GeoFormer processes fused multi-source remote sensing data---namely Sentinel-1 (SAR), Sentinel-2 (optical), and DEM---into a unified 8-band multi-channel tensor (see Fig.~\ref{fig:input_tensor}). An additional binary mask channel is included to indicate valid (non-padded) regions, enabling the model to distinguish between central and contextual patches during both training and inference. This mask serves as a spatial prior that guides attention and loss computation toward valid areas, particularly when using dynamically constructed context windows.

The input tensor is then divided into fixed-size non-overlapping patches (e.g., $3\times3$, $5\times5$, or $9\times9$ context windows), and each patch is linearly embedded into a token sequence as in the original Swin Transformer~\cite{liu2021swin}.

The token sequence passes through two successive Swin Transformer blocks, each consisting of alternating window-based multi-head self-attention (W-MSA) and shifted window attention (SW-MSA) modules, layer normalization (LN), and feedforward multilayer perceptrons (MLPs). This design allows GeoFormer to capture both local and global spatial dependencies efficiently.

After the backbone processing, we apply an \textbf{extract-center} module to isolate the representation of the central patch from the surrounding context. This center token---representing the target $100\,\mathrm{m}$ grid---is then passed through an average pooling layer and fed into two independent prediction heads.

Each head is a lightweight MLP tailored to one of the two tasks:
\begin{itemize}
  \item The BH head uses a ReLU activation to regress building height.
  \item The BF head uses a sigmoid activation to constrain the footprint ratio between 0 and 1.
\end{itemize}

This architecture provides spatially aligned multi-task outputs, ensures computational efficiency via hierarchical feature reuse, and allows flexible control over the receptive field size through the patch grid configuration.

\subsection{Loss Function}

To jointly predict building height (\( y^{\mathrm{bh}} \)) and building footprint ratio (\( y^{\mathrm{bf}} \)), this study employs a multi-task loss function composed of two regression objectives. Following uncertainty-based weighting~\cite{kendall2018multi}, the total loss is defined as:

\begin{equation}
\mathcal{L}_{\text{total}} = \frac{1}{2\sigma_{\mathrm{bh}}^2} \mathcal{L}_{\mathrm{bh}} + \frac{1}{2\sigma_{\mathrm{bf}}^2} \mathcal{L}_{\mathrm{bf}} + \log \sigma_{\mathrm{bh}} + \log \sigma_{\mathrm{bf}}
\end{equation}

Here, \( \mathcal{L}_{\mathrm{bh}} \) and \( \mathcal{L}_{\mathrm{bf}} \) are regression losses for building height and footprint respectively, and \( \sigma_{\mathrm{bh}} \), \( \sigma_{\mathrm{bf}} \) are learnable task uncertainties optimized during training.

For each task, the Adaptive Huber Loss~\cite{huber1964robust} is used, which combines the sensitivity of Mean Squared Error (MSE) with the robustness of Mean Absolute Error (MAE). It is defined as:

\begin{equation}
\mathrm{HuberLoss}(\hat{y}, y) =
\begin{cases}
\frac{1}{2\delta} (\hat{y} - y)^2, & \text{if } |\hat{y} - y| < \delta \\
|\hat{y} - y| - \frac{\delta}{2}, & \text{otherwise}
\end{cases}
\end{equation}

Here, \( \hat{y} \) is the predicted value, \( y \) is the ground-truth, and \( \delta \) controls the transition between the quadratic and linear regimes. When residuals are small, the loss behaves like MSE; for large residuals, it behaves like MAE to mitigate the influence of outliers.

At the beginning of training, \( \delta \) is initialized based on the Mean Absolute Error (MAE) over the training set:

\begin{equation}
\delta^{(0)} = \frac{1}{N} \sum_{j=1}^{N} |\hat{y}_j - y_j|
\end{equation}

During training, \( \delta \) is updated at each epoch using the residual statistics of the current model, allowing the loss function to dynamically adapt to prediction uncertainty.

This hybrid and adaptive design ensures both robustness and precision in geospatial regression tasks, making it well-suited for predicting building properties from satellite imagery.

\subsection{Optimization and Training Settings}

The model is trained using the AdamW optimizer~\cite{loshchilov2017decoupled}, with an initial learning rate of $1\times10^{-4}$ and a weight decay of $1\times10^{-2}$. To stabilize early-stage training and encourage smooth convergence, we adopt a cosine annealing learning rate schedule with warmup~\cite{loshchilov2016sgdr}.

Mixed-precision training is enabled using automatic mixed precision (AMP) to accelerate convergence and reduce GPU memory usage. Early stopping and checkpointing are implemented based on validation $R^2$ to ensure optimal generalization.

All experiments are conducted with a batch size of 64 and trained for up to 150 epochs or until convergence.

\section{Model Evaluation}

\subsection{Evaluation Metrics and Justification}

\begin{table*}[!t]
  \centering
  \caption{Evaluation metrics used for regression accuracy assessment}
  \label{tab:metrics}
  \renewcommand{\arraystretch}{1.3}
  \begin{tabular}{@{}p{0.10\textwidth} p{0.34\textwidth} p{0.12\textwidth} p{0.12\textwidth} p{0.20\textwidth}@{}}
    \toprule
    \textbf{Metric} & \textbf{Formula} & \textbf{Range} & \textbf{Direction} & \textbf{Focus} \\
    \midrule
    RMSE & \( \sqrt{\frac{1}{N} \sum_{i=1}^{N} (\hat{y}_i - y_i)^2} \) & $[0,\infty)$ & Lower & Sensitive to outliers \\
    MAE & \( \frac{1}{N} \sum_{i=1}^{N} |\hat{y}_i - y_i| \) & $[0,\infty)$ & Lower & Average error magnitude \\
    ME  & \( \frac{1}{N} \sum_{i=1}^{N} (\hat{y}_i - y_i) \) & $(-\infty,\infty)$ & Close to 0 & Overall bias / signed error \\
    NMAD & \( 1.4826 \times \mathrm{median} \left( |\hat{y}_i - y_i - \mathrm{median}(\hat{y}_i - y_i)| \right) \) & $[0,\infty)$ & Lower & Robust to outliers \\
    CC & \( \frac{\mathrm{Cov}(\hat{y}, y)}{\sigma_{\hat{y}} \sigma_y} \) & $[-1, 1]$ & Close to 1 & Linear correlation \\
    \( R^2 \) & \( 1 - \frac{\sum_{i=1}^{N} (\hat{y}_i - y_i)^2}{\sum_{i=1}^{N} (y_i - \bar{y})^2} \) & $(-\infty, 1]$ & Close to 1 & Variance explained by model \\
    \bottomrule
  \end{tabular}
\end{table*}

Table~\ref{tab:metrics} summarizes the set of regression metrics employed in this study to evaluate the predictive performance of building height and footprint estimation. These metrics are carefully selected to provide a multi-perspective assessment, covering magnitude-based errors (RMSE, MAE), bias (ME), robustness to outliers (NMAD), and statistical correlation with ground-truth labels (CC and \(R^2\)). 

Root Mean Squared Error (RMSE) penalizes large residuals more heavily, making it sensitive to outliers, while Mean Absolute Error (MAE) captures the average prediction error in a more balanced manner. Mean Error (ME) provides a signed indication of systematic bias in predictions. The Normalized Median Absolute Deviation (NMAD) is particularly suited for datasets with non-Gaussian error distributions, offering a robust alternative to RMSE. Pearson's Correlation Coefficient (CC) and the Coefficient of Determination (\(R^2\)) jointly evaluate the consistency and explanatory power of the model outputs relative to the reference values. 

Together, these metrics allow for a comprehensive evaluation of both overall model accuracy and stability under heterogeneous spatial conditions.

\subsection{CNN Baseline Comparison}

Before evaluating GeoFormer, we first benchmark three representative CNN architectures as single-patch baselines. Each model receives a single $10 \times 10$ pixel patch (corresponding to one $100\,\mathrm{m}$ grid cell) and outputs joint BH and BF predictions through a shared backbone with two task-specific heads. All baselines adopt the same multi-task Huber loss and uncertainty-based weighting described in Section~III-B, and are trained under identical data splits, augmentation, and optimization settings. The three architectures are:

\begin{itemize}
  \item \textbf{ResNet-MTL}~\cite{he2016deep}: a residual network with skip connections that alleviate vanishing gradients and enable deeper feature hierarchies.
  \item \textbf{UNet-MTL}~\cite{ronneberger2015u}: an encoder--decoder network with lateral skip connections that preserve fine-grained spatial details through multi-scale feature fusion.
  \item \textbf{SENet-MTL}~\cite{hu2018squeeze}: a squeeze-and-excitation network that recalibrates channel-wise feature responses via learned attention weights, building upon the architecture used in the original SHAFTS framework~\cite{shafts2023}.
\end{itemize}

\begin{figure}[!t]
  \centering
  \includegraphics[width=\linewidth]{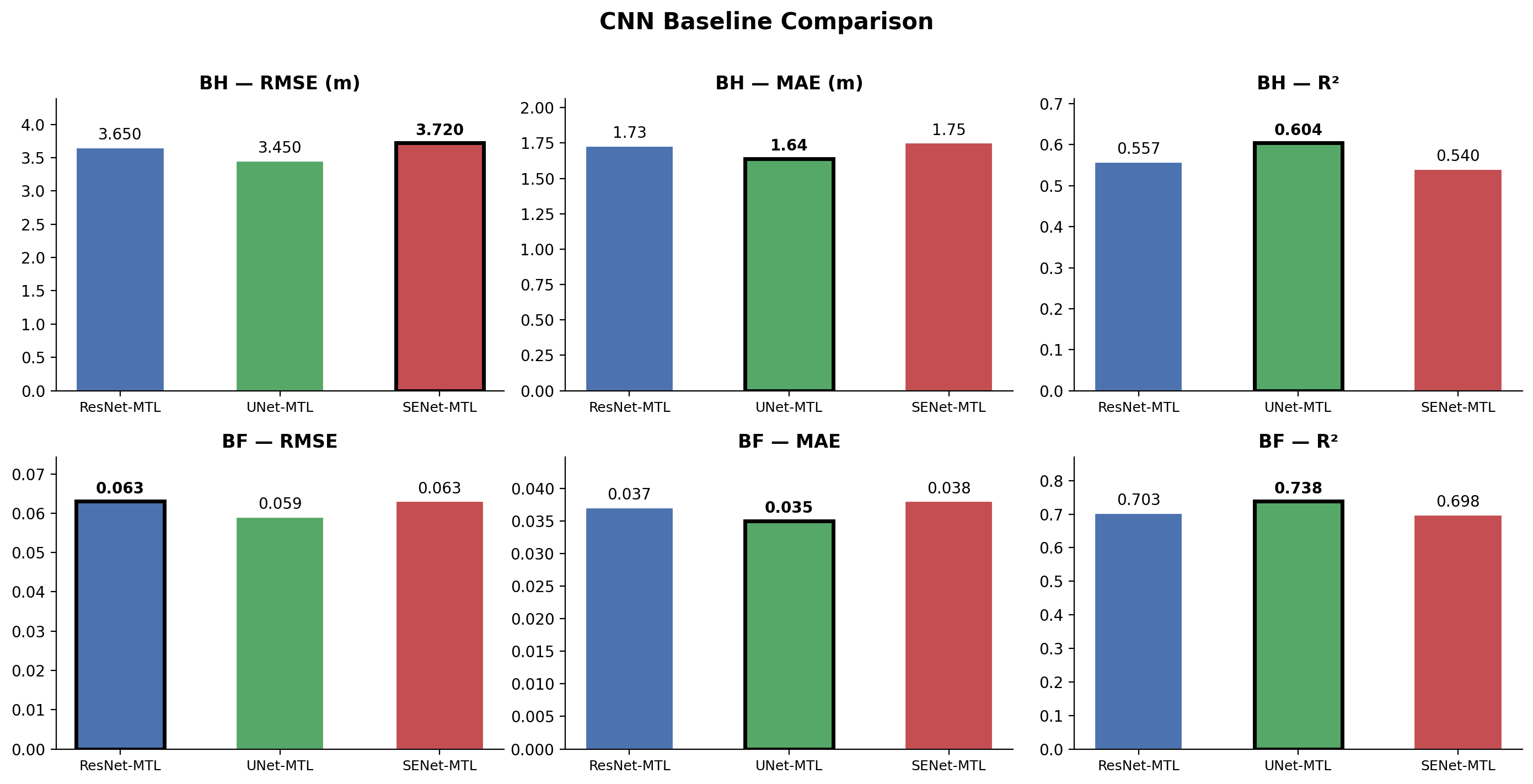}
  \caption{Comparison of three CNN baseline architectures on BH and BF prediction. UNet-MTL achieves the best overall performance across all metrics.}
  \label{fig:baseline_comparison}
\end{figure}

As shown in Fig.~\ref{fig:baseline_comparison}, UNet-MTL consistently achieves the best performance among the three CNN baselines for both BH and BF tasks. Specifically, UNet-MTL yields BH RMSE of 3.45~m and $R^2$ of 0.604, outperforming ResNet-MTL by 5.5\% and SENet-MTL by 7.3\% in RMSE. For BF prediction, UNet-MTL achieves the highest $R^2$ (0.738) with the lowest RMSE (0.059). The performance ranking---UNet $>$ ResNet $>$ SENet---is consistent across both tasks, suggesting that the encoder--decoder feature pyramid in UNet provides a stronger inductive bias for dense spatial regression than either residual depth (ResNet) or channel attention alone (SENet).

Based on these results, UNet-MTL is selected as the strongest CNN baseline and is carried forward into all subsequent comparisons with GeoFormer.

\subsection{Experimental Results}

To validate the effectiveness of the proposed \textbf{GeoFormer} framework, we compare it with the baseline model \textbf{UNet-MTL}, which was selected as the strongest CNN baseline in the preceding comparison. A detailed description of this model can be found in the Appendix.

\begin{figure*}[!t]
\centering
\includegraphics[width=\linewidth]{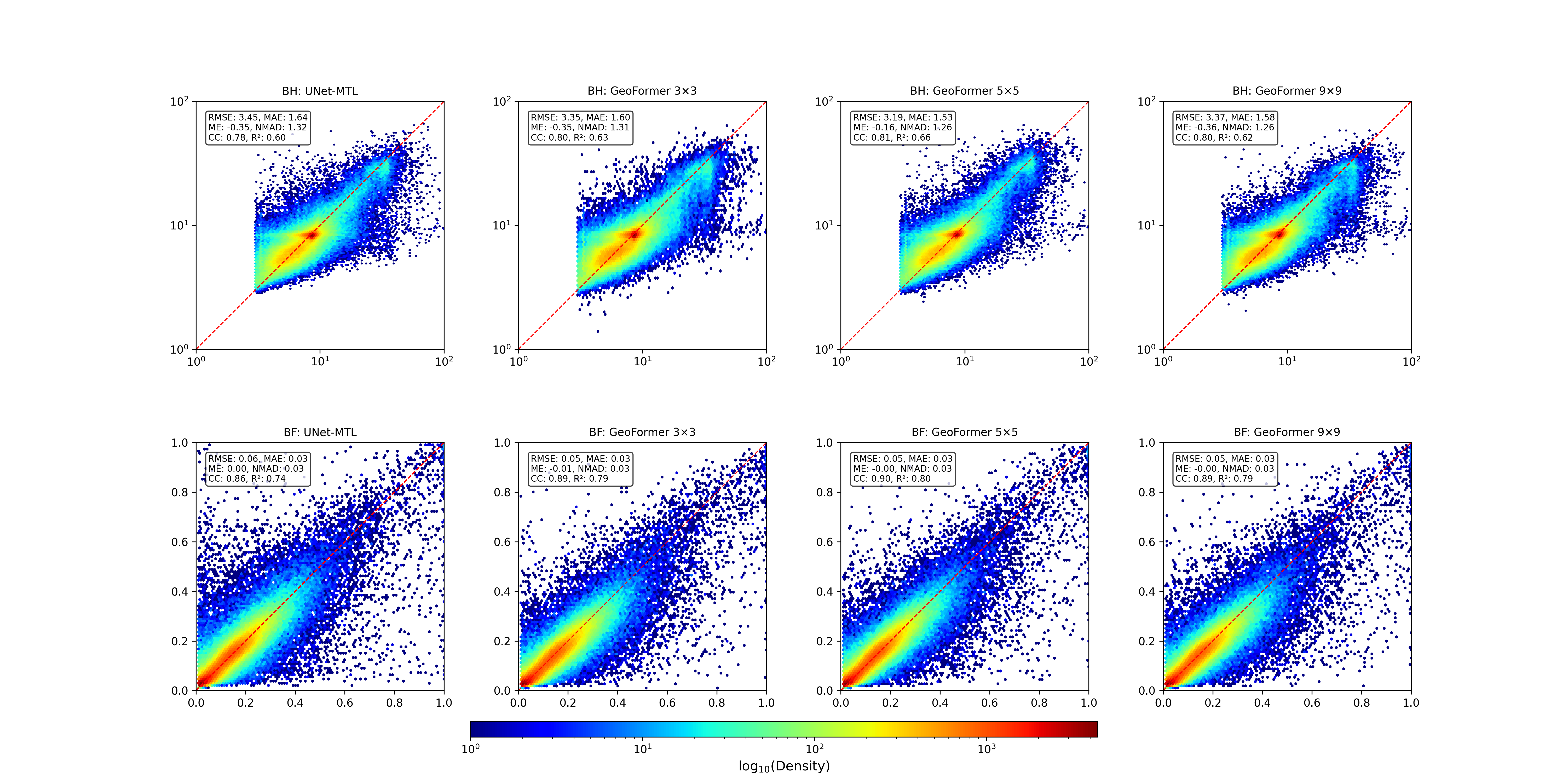}
\caption{Scatter plots comparing predictions and ground truths for BH and BF under different receptive field configurations.}
\label{fig:comparison}
\end{figure*}

As shown in Fig.~\ref{fig:comparison}, the predicted values from GeoFormer are more tightly clustered along the identity line compared to the baseline UNet-MTL, indicating higher accuracy and reduced systematic bias. This effect is especially noticeable for the \(5 \times 5\) receptive field, which shows the densest distribution around the diagonal in both building height (BH, top row) and building footprint (BF, bottom row) panels. 

In the BH panels, the baseline UNet-MTL tends to systematically underestimate taller buildings, particularly in the high-value range (\(> 30~\mathrm{m}\)), as evidenced by the visible skew below the red dashed line. GeoFormer significantly mitigates this bias, especially in the 3×3 and 5×5 configurations. The 5×5 variant yields the lowest residual dispersion and the most symmetric pattern, suggesting superior generalization and robustness.

In the BF panels (bottom row), all GeoFormer variants produce denser clusters closer to the identity line, with the 5×5 configuration again showing the strongest alignment. The color scale, indicating the logarithm of point density, confirms this improvement through brighter concentrations along the diagonal. This suggests not only improved accuracy but also fewer large outliers in footprint prediction.

Quantitative metrics that support these observations are presented in Table~\ref{tab:eval_metrics}.

\begin{table}[!t]
\centering
\small
\setlength{\tabcolsep}{3pt}
\caption{Quantitative evaluation of BH and BF predictions across models and receptive fields.}
\label{tab:eval_metrics}
\begin{tabular}{lcccccc}
\toprule
\textbf{Model} & \textbf{RMSE} & \textbf{MAE} & \textbf{ME} & \textbf{NMAD} & \textbf{CC} & \boldmath$R^2$ \\
\midrule
\multicolumn{7}{c}{\textit{Building Height (BH)}} \\
UNet-MTL        & 3.45 & 1.64 & $-$0.35 & 1.32 & 0.78 & 0.60 \\
GeoFormer 3$\times$3   & 3.35 & 1.60 & $-$0.35 & 1.31 & 0.80 & 0.63 \\
GeoFormer 5$\times$5   & \textbf{3.19} & \textbf{1.53} & \textbf{$-$0.16} & \textbf{1.26} & \textbf{0.81} & \textbf{0.66} \\
GeoFormer 9$\times$9   & 3.37 & 1.58 & $-$0.36 & 1.26 & 0.80 & 0.62 \\
\midrule
\multicolumn{7}{c}{\textit{Building Footprint (BF)}} \\
UNet-MTL        & 0.059 & 0.034 & 0.000 & 0.034 & 0.86 & 0.74 \\
GeoFormer 3$\times$3   & 0.053 & 0.032 & $-$0.005 & 0.031 & 0.89 & 0.79 \\
GeoFormer 5$\times$5   & \textbf{0.050} & \textbf{0.031} & \textbf{0.000} & \textbf{0.030} & \textbf{0.90} & \textbf{0.80} \\
GeoFormer 9$\times$9   & 0.052 & 0.032 & 0.000 & 0.031 & 0.89 & 0.79 \\
\bottomrule
\end{tabular}
\end{table}

Table~\ref{tab:eval_metrics} summarizes the quantitative evaluation results for both building height (BH) and building footprint (BF) prediction across all architectures tested. Several important findings emerge from this comparison.

For BH prediction, GeoFormer with a \(5 \times 5\) receptive field achieves the lowest RMSE (3.19\,m) and MAE (1.53\,m), representing a relative reduction of 7.5\,\% and 6.7\,\%, respectively, compared to the best CNN baseline UNet-MTL (RMSE: 3.45\,m, MAE: 1.64\,m). The mean error (ME) is significantly closer to zero ($-$0.16 vs.\ $-$0.35), indicating reduced systematic bias, and the $R^2$ rises from 0.60 to 0.66.

Overall, the \(5 \times 5\) receptive field configuration offers the most balanced trade-off between accuracy, robustness, and generalization. Enlarging the receptive field to \(9 \times 9\) does not yield further gains and slightly degrades BH performance (RMSE 3.37\,m), likely due to over-smoothing from excessive contextual aggregation.

\begin{table}[!t]
\centering
\caption{Computational efficiency comparison. Params = trainable parameters; FLOPs measured on a single $8 \times 50 \times 50$ input; inference time measured on an NVIDIA RTX 3090 GPU with batch size 1 (mean of 200 runs after 50 warm-up iterations).}
\label{tab:efficiency}
\begin{tabular}{lccc}
\toprule
\textbf{Model} & \textbf{Params (M)} & \textbf{FLOPs (M)} & \textbf{Time (ms)} \\
\midrule
ResNet-18       & 11.19 & 133.7 & 1.34 \\
SE-ResNet-18    & 11.28 & 133.8 & 2.08 \\
UNet-Enc        &  4.94 & 498.4 & 0.74 \\
ConvNeXt-T      & 27.83 & 187.0 & 2.97 \\
\midrule
GeoFormer (base)  &  \textbf{0.32} & 148.0 & 1.05 \\
GeoFormer (large) &  2.37 & 582.9 & 1.96 \\
\bottomrule
\end{tabular}
\end{table}

Table~\ref{tab:efficiency} reports the computational cost of each model. GeoFormer (base) contains only 0.32\,M trainable parameters---roughly 35$\times$ fewer than ResNet-18 and 87$\times$ fewer than ConvNeXt-T---while maintaining competitive FLOPs and inference latency. Even the enlarged GeoFormer (large) variant (2.37\,M) remains lighter than all other baselines except UNet-Enc. These results demonstrate that the Swin Transformer backbone, combined with center-crop pooling, yields a highly parameter-efficient architecture suitable for large-scale global mapping where computational budget is a practical constraint.

\subsection{Error Source Analysis}

To complement the quantitative evaluation, we perform an in-depth analysis of prediction errors across different building characteristics. Residuals are stratified by height ($H_{\mathrm{ave}}$) and footprint density ($\lambda_p$) to reveal patterns linked to sample rarity, spatial imbalance, and model-specific behavior.

\begin{figure}[!t]
  \centering
  \includegraphics[width=\linewidth]{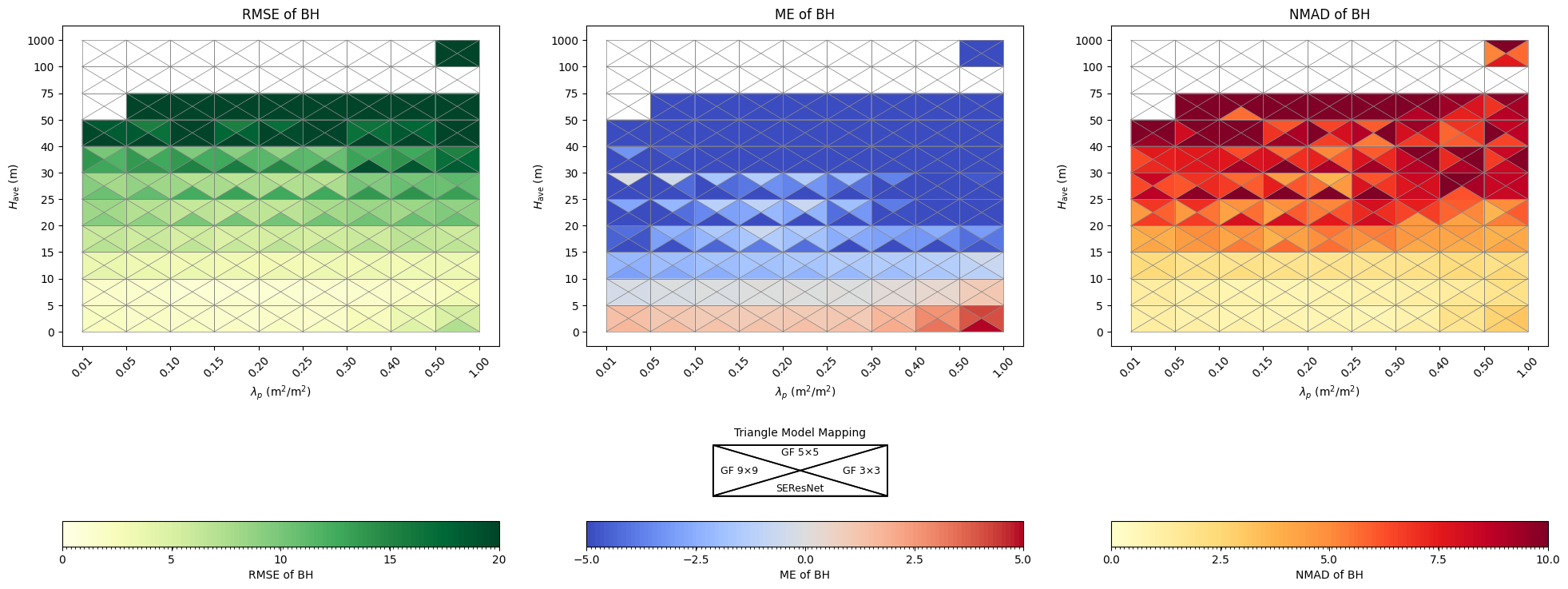}
  \caption{Stratified error analysis of building height prediction across height and density bins}
  \label{fig:bh_error_source}
\end{figure}

Fig.~\ref{fig:bh_error_source} presents the stratified error analysis for building height (BH) predictions, plotted over binned grids of average building height ($H_{\mathrm{ave}}$) and footprint density ($\lambda_p$). The three panels respectively show the root mean square error (RMSE), mean error (ME), and normalized median absolute deviation (NMAD). A consistent trend is observed across all models: in regions where $H_{\mathrm{ave}} > 50$~m, both RMSE and NMAD increase substantially, highlighting the inherent difficulty of predicting rare high-rise buildings.

Furthermore, the ME panel reveals that UNet-MTL and GeoFormer 9$\times$9 tend to exhibit stronger negative bias---i.e., underestimation---especially in these tall-building regions. This underperformance is likely due to limited representation of such samples in the training data. In contrast, the GeoFormer variants with 5$\times$5 and 3$\times$3 receptive fields demonstrate better calibration, with lower errors and reduced bias. These results collectively indicate that smaller receptive fields help preserve critical spatial details and improve robustness when dealing with vertical complexity.

\begin{figure}[!t]
  \centering
  \includegraphics[width=\linewidth]{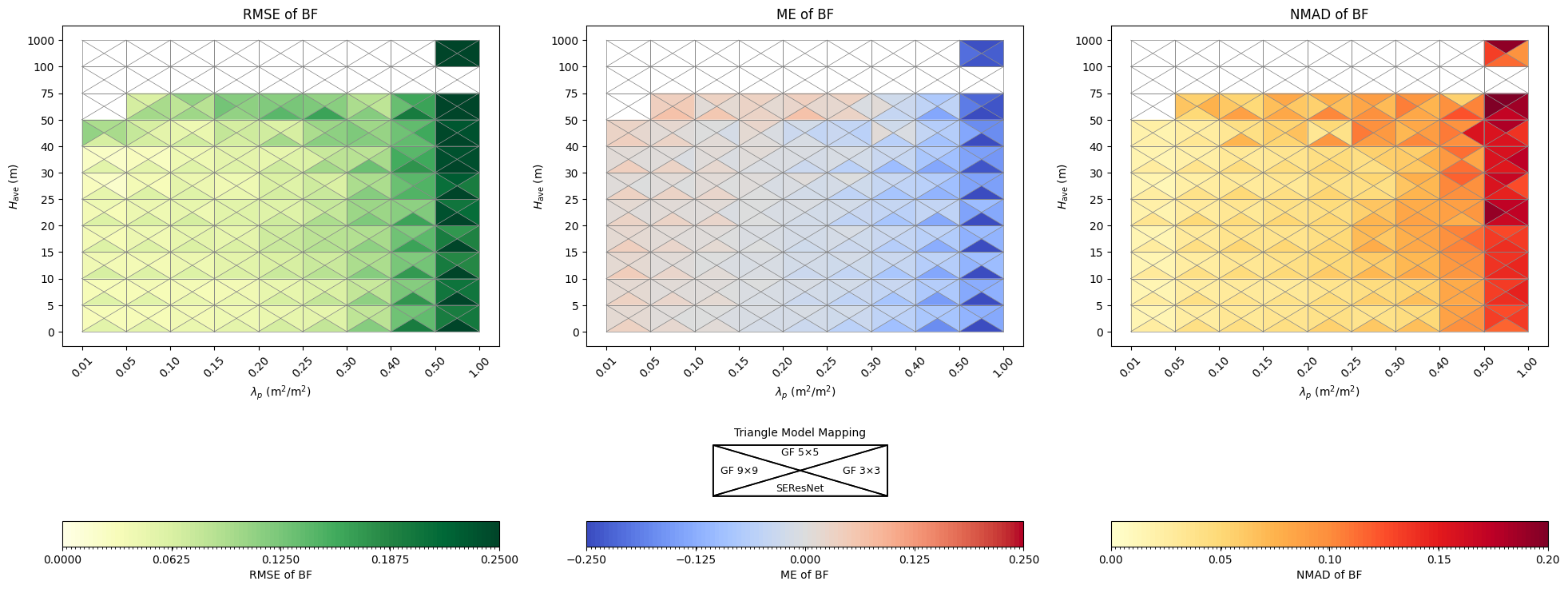}
  \caption{Stratified error analysis of building footprint prediction across height and density bins}
  \label{fig:bf_error_source}
\end{figure}

Fig.~\ref{fig:bf_error_source} illustrates the stratified error maps for building footprint (BF) predictions across the same binned space of average building height ($H_{\mathrm{ave}}$) and footprint density ($\lambda_p$). A notable pattern emerges: prediction errors, particularly RMSE and NMAD, increase markedly in regions with high planar density ($\lambda_p > 0.5$), indicating that all models face difficulty in densely built environments.

The ME plot further reveals a consistent underestimation trend across models in these high-density regions. This systematic negative bias may be attributed to the scarcity of compact patches in the training dataset, which leads to poorer generalization in these cases. Among the evaluated models, GeoFormer with a 5$\times$5 receptive field exhibits the most stable and balanced performance, maintaining lower errors and reduced bias even under extreme density conditions. This suggests that moderate contextual aggregation improves footprint estimation while avoiding the over-smoothing observed in larger receptive fields.

\begin{figure}[!t]
  \centering
  \includegraphics[width=\linewidth]{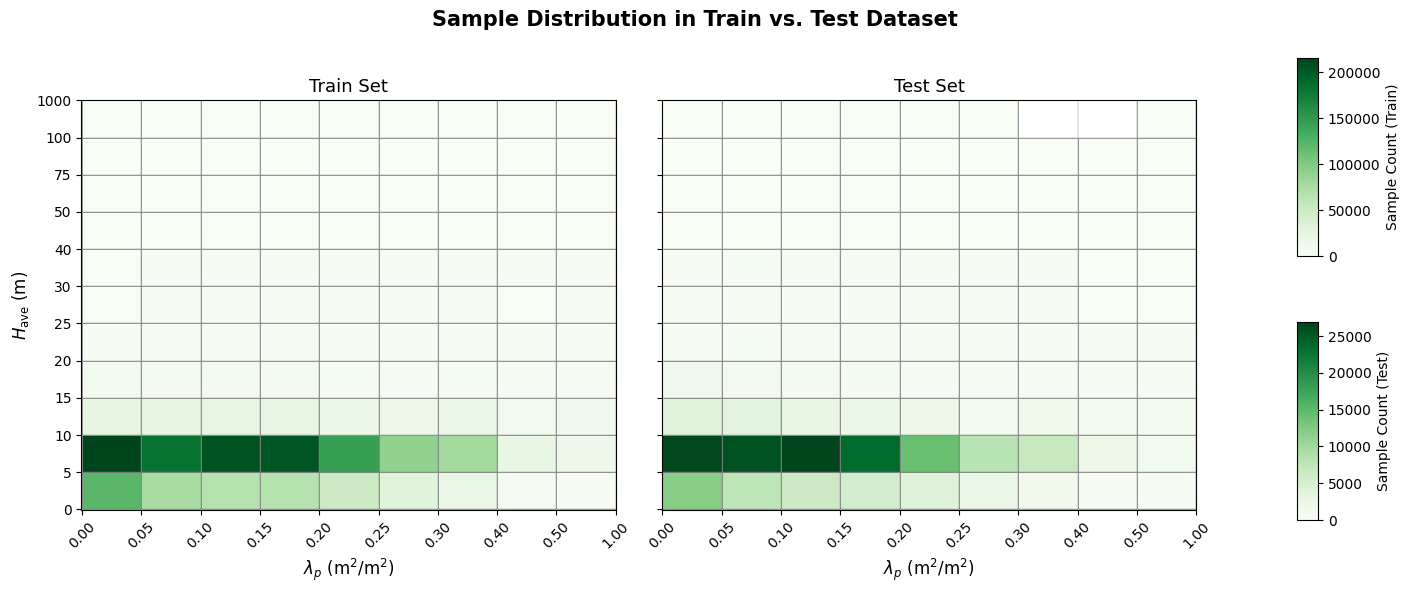}
  \caption{Sample distribution in train vs. test dataset.}
  \label{fig:sample-distribution}
\end{figure}

Fig.~\ref{fig:sample-distribution} illustrates the joint distribution of building height ($H_{\mathrm{ave}}$) and footprint ratio ($\lambda_p$) in the train and test datasets. A pronounced long-tailed pattern is observed, where samples with $H_{\mathrm{ave}} > 30$~m and $\lambda_p > 0.7$ are extremely rare, accounting for only 1.1\% and 0.54\% of the total data, respectively. This imbalance explains the model's degraded performance in high-rise and dense areas, as the network receives insufficient training signals from these underrepresented regimes. In fact, the model performs robustly on the majority of cases, and the main bottleneck lies not in architectural flaws, but in data coverage. Future efforts should prioritize collecting more representative samples from vertical and compact urban environments to improve generalization.

\begin{figure}[!t]
    \centering
    \includegraphics[width=0.85\linewidth]{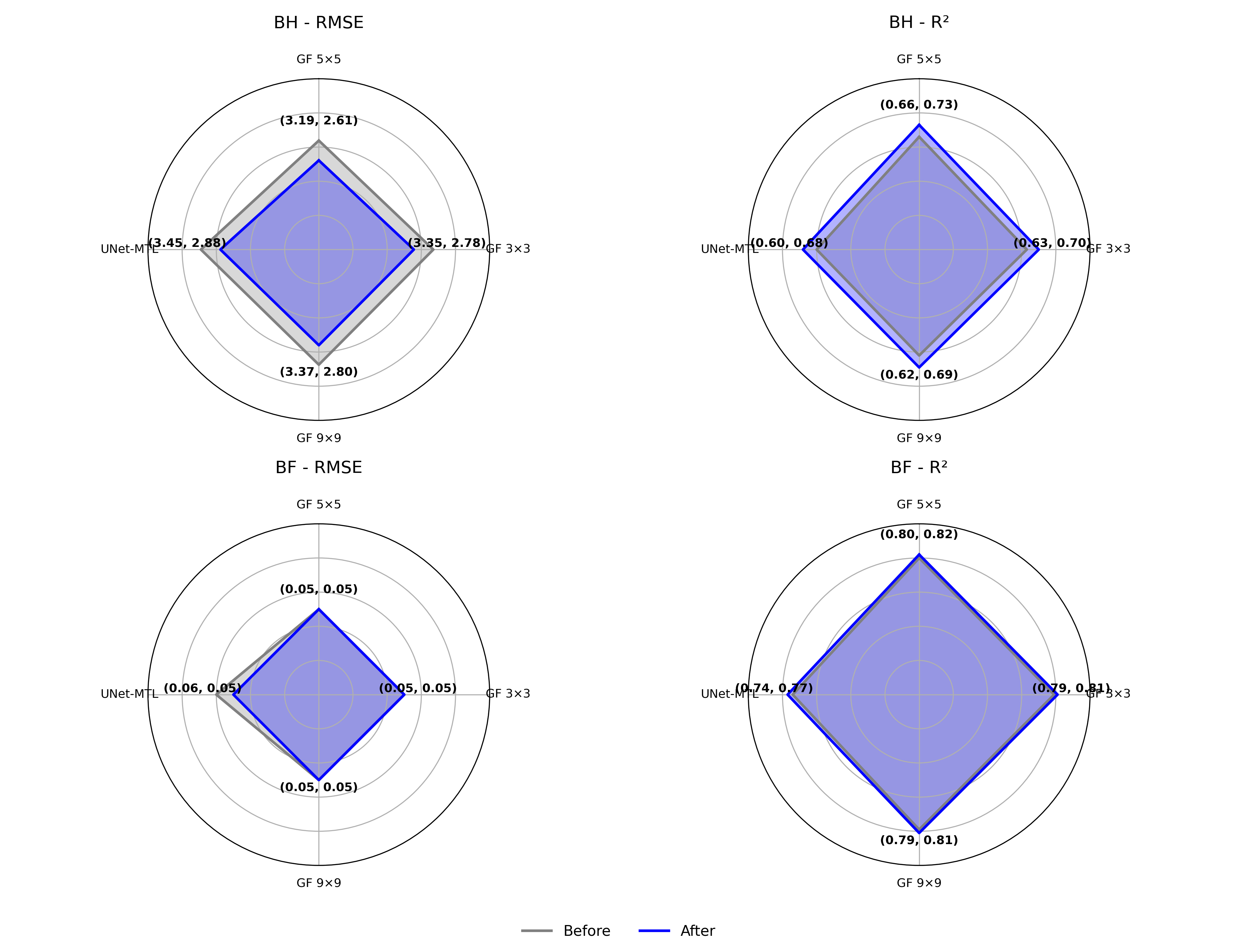}
    \caption{Performance before and after removing the top 0.1\% residuals across four model configurations for BH and BF estimation.}
    \label{fig:radar_outlier_removed}
\end{figure}

To mitigate the impact of extreme prediction errors, we further removed the top 0.1\% of residuals from each model's output. As shown in Fig.~\ref{fig:radar_outlier_removed}, this adjustment led to consistent improvements in both RMSE and $R^2$ metrics across all model configurations. The enhancement was particularly notable for building height (BH) prediction, indicating that tall buildings often constitute the most outlying errors in the residual distribution.

Overall, the GeoFormer series exhibited more stable and consistent improvements after outlier removal, especially in configurations with larger receptive fields (5×5 and 9×9). This suggests that incorporating broader spatial context plays a critical role in suppressing localized prediction anomalies.

In summary, removing extreme residuals helps reveal the core predictive capacity of each model and highlights the importance of model structure in ensuring robustness.

\section{Ablation Study}
\label{sec:ablation}

To systematically evaluate the design choices underlying GeoFormer, we conduct four ablation experiments organised into two groups: \textbf{structural ablation} (model capacity and DEM removal) and \textbf{modality ablation} (SAR and optical channel removal). All ablation variants share the same training protocol and GeoSplit partitioning as the full GeoFormer (5$\times$5) model, differing only in the factor under investigation.

\subsection{Structural Ablation}

\subsubsection{Effect of Model Capacity (Enlarged Model)}
\label{sec:ablation-enlarged}

A natural hypothesis is that increasing model capacity should improve prediction accuracy. To test this, we train an \emph{Enlarged} GeoFormer variant in which the Swin Transformer embedding dimension and depth are doubled relative to the baseline configuration, roughly quadrupling the number of trainable parameters. Table~\ref{tab:ablation-structural} shows the results on the three-city held-out test set.

\begin{figure}[!t]
\centering
\includegraphics[width=\linewidth]{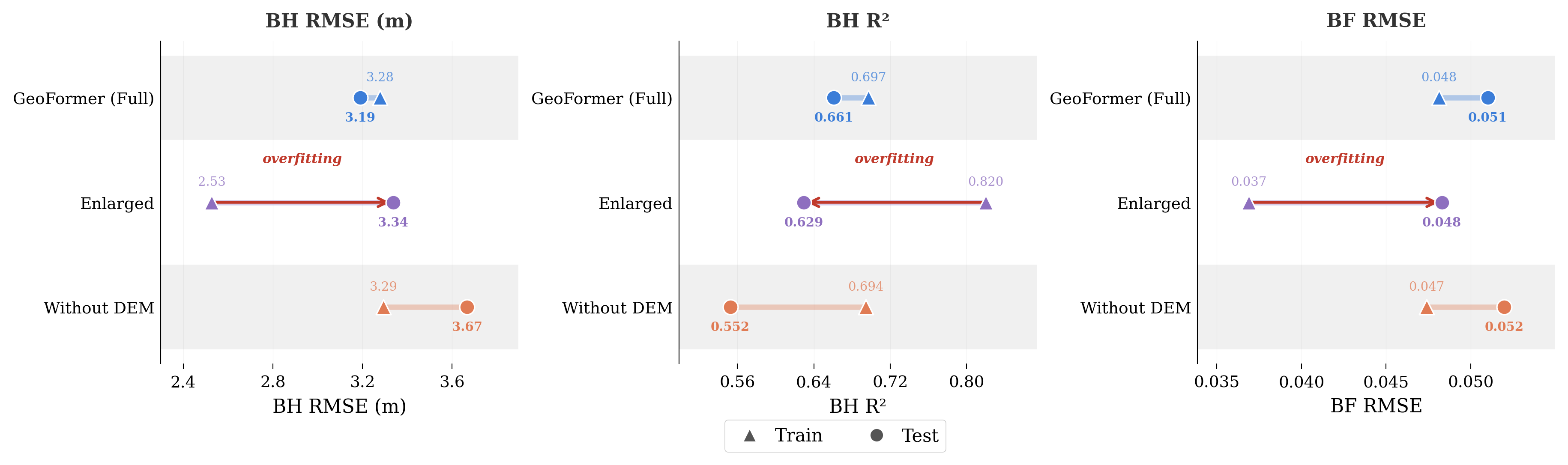}
\caption{Train vs.\ test performance gap for GeoFormer (Full), Enlarged, and Without-DEM variants. The Enlarged model exhibits a clear overfitting pattern: substantially lower training error but higher test error.}
\label{fig:ablation-train-test-gap}
\end{figure}

The Enlarged model achieves noticeably lower training error (BH RMSE 2.53\,m vs.\ 3.28\,m for the baseline), yet its test-set BH RMSE increases to 3.34\,m compared with 3.19\,m for the standard model, and the BH R\textsuperscript{2} drops from 0.661 to 0.629. As illustrated in Fig.~\ref{fig:ablation-train-test-gap}, the widening train--test gap in the Enlarged model constitutes clear evidence of overfitting: the extra parameters memorise training-set specifics rather than learning generalisable urban morphology features. This finding justifies the compact architecture adopted for the final GeoFormer model.

\subsubsection{Effect of Removing the DEM Channel}
\label{sec:ablation-dem}

DEM encodes local terrain elevation, which is conceptually correlated with building height. To quantify its contribution, we retrain GeoFormer with DEM removed from the input stack, retaining only SAR and optical channels. As shown in Table~\ref{tab:ablation-structural}, removing DEM leads to the largest BH degradation among all ablation variants: BH RMSE rises from 3.19\,m to 3.67\,m (+15.0\,\%) and BH R\textsuperscript{2} falls from 0.661 to 0.552 ($-$16.5\,\%). In contrast, BF RMSE increases only marginally (0.051 to 0.052).

This asymmetric impact confirms that DEM provides essential vertical-scale cues for height estimation but contributes little to planar footprint prediction, where spectral and textural information already suffice.

\begin{table}[!t]
\centering
\small
\setlength{\tabcolsep}{3pt}
\caption{Structural ablation results on the three-city held-out test set.}
\label{tab:ablation-structural}
\begin{tabular}{l ccc ccc}
\toprule
 & \multicolumn{3}{c}{\textbf{Building Height}} & \multicolumn{3}{c}{\textbf{Building Footprint}} \\
\cmidrule(lr){2-4} \cmidrule(lr){5-7}
\textbf{Model} & MAE & RMSE & R\textsuperscript{2} & MAE & RMSE & R\textsuperscript{2} \\
\midrule
GeoFormer (Full) & 1.44 & 3.19 & 0.661 & 0.031 & 0.051 & 0.801 \\
Enlarged          & 1.52 & 3.34 & 0.629 & 0.029 & 0.048 & 0.823 \\
Without DEM       & 1.68 & 3.67 & 0.552 & 0.032 & 0.052 & 0.794 \\
\bottomrule
\end{tabular}
\end{table}

\subsection{Modality Ablation}

\subsubsection{Effect of Removing the SAR Channel (No-SAR)}
\label{sec:ablation-nosar}

To assess the contribution of SAR backscatter, we train GeoFormer with only Sentinel-2 optical and DEM inputs. As reported in Table~\ref{tab:ablation-modality}, removing SAR incurs a moderate degradation in BH performance (RMSE: 3.19 $\rightarrow$ 3.38\,m, $+$6.0\,\%), while BF RMSE shows a somewhat larger relative increase (0.051 $\rightarrow$ 0.059, $+$15.7\,\%). This suggests that SAR's primary contribution lies in capturing three-dimensional scattering features related to structural density, which are not fully captured by optical reflectance alone.

\subsubsection{Effect of Removing the Optical Channel (No-Optical)}
\label{sec:ablation-nooptical}

\begin{figure}[!t]
\centering
\includegraphics[width=\linewidth]{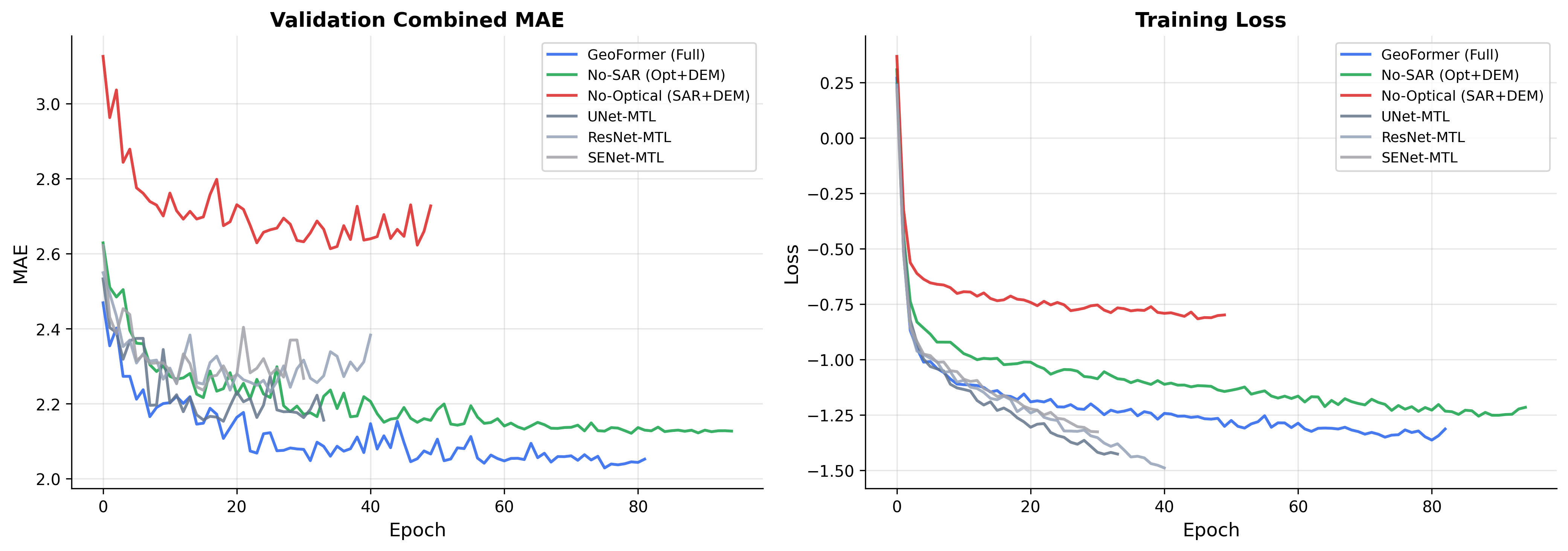}
\caption{Validation combined MAE and training loss curves for GeoFormer variants and CNN baselines. All runs were configured for 150 epochs but terminated earlier by the early-stopping mechanism (patience\,=\,10 on validation R\textsuperscript{2}), hence the varying curve lengths. The No-Optical configuration consistently exhibits higher error, while the full model and No-SAR variant converge to similar levels.}
\label{fig:ablation-training-curves}
\end{figure}

Conversely, removing Sentinel-2 optical data and retaining only SAR and DEM results in a dramatically larger performance drop. BH RMSE increases to 4.40\,m ($+$37.9\,\%), BH R\textsuperscript{2} collapses from 0.661 to 0.357, and BF RMSE rises to 0.074 ($+$45.1\,\%). The No-Optical model consistently exhibits the highest validation MAE throughout training (Fig.~\ref{fig:ablation-training-curves}), confirming that multispectral reflectance is the dominant information source for both height and footprint retrieval at 100\,m resolution.

\begin{table}[!t]
\centering
\footnotesize
\setlength{\tabcolsep}{2pt}
\caption{Modality ablation results evaluated on the 51-city global test set (Test2).}
\label{tab:ablation-modality}
\begin{tabular}{l ccc ccc}
\toprule
 & \multicolumn{3}{c}{\textbf{Building Height}} & \multicolumn{3}{c}{\textbf{Building Footprint}} \\
\cmidrule(lr){2-4} \cmidrule(lr){5-7}
\textbf{Config.} & MAE & RMSE & R\textsuperscript{2} & MAE & RMSE & R\textsuperscript{2} \\
\midrule
GeoFormer (Full)       & 1.53 & 3.19 & 0.661 & 0.031 & 0.051 & 0.801 \\
No-SAR (Opt+DEM)       & 1.62 & 3.38 & 0.620 & 0.036 & 0.059 & 0.735 \\
No-Optical (SAR+DEM)   & 2.14 & 4.40 & 0.357 & 0.047 & 0.074 & 0.581 \\
\bottomrule
\end{tabular}
\end{table}

\subsection{Summary of Ablation Findings}

\begin{figure}[!t]
\centering
\includegraphics[width=\linewidth]{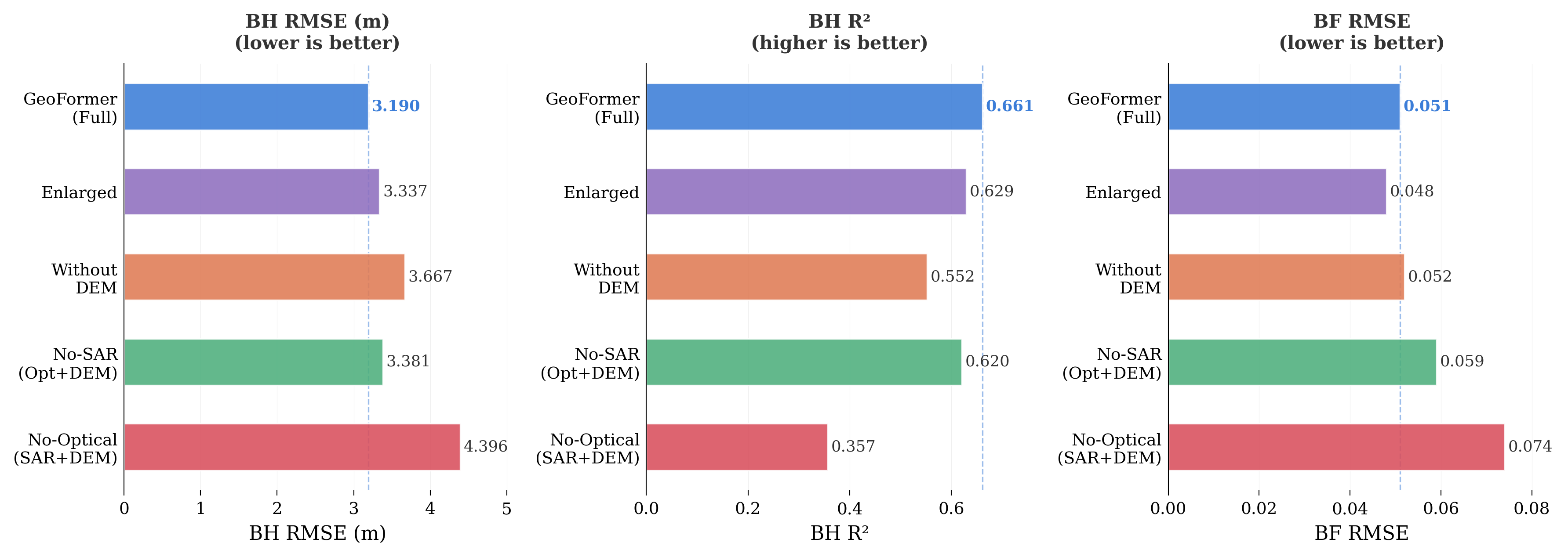}
\caption{Consolidated comparison of all ablation variants. Top row: structural ablation (capacity and DEM); bottom row: modality ablation (SAR and optical).}
\label{fig:ablation-summary}
\end{figure}

Fig.~\ref{fig:ablation-summary} consolidates the results from all four ablation experiments. Three principal conclusions emerge:

\begin{enumerate}
    \item \textbf{Model capacity must be carefully calibrated.} The Enlarged variant achieves lower training error but higher test error, demonstrating that naively scaling model parameters leads to overfitting rather than improved generalisation in this data regime.
    \item \textbf{DEM is indispensable for height estimation.} Removing DEM causes the largest single-factor degradation in BH metrics ($+$15.0\,\% RMSE, $-$16.5\,\% R\textsuperscript{2}) while barely affecting BF, confirming that topographic context provides unique vertical-scale information not recoverable from spectral data alone.
    \item \textbf{Optical imagery is the dominant modality.} The No-Optical configuration suffers the most severe performance collapse across all metrics ($+$37.9\,\% BH RMSE). By contrast, removing SAR causes only moderate degradation ($+$6.0\,\% BH RMSE), indicating that multispectral reflectance carries the primary predictive signal at 100\,m resolution. Nevertheless, the full SAR+Optical+DEM model remains the top performer, validating the complementary value of multi-source fusion.
\end{enumerate}

\section{Evaluating Model Generalization}

\subsection{Cross-Dataset Generalization: Test on Suwon, Korea}

To evaluate GeoFormer's generalization beyond the training domain, we test its performance on Suwon, South Korea---a large East Asian city with distinct morphology and sensor conditions. Unlike the high-rise, hyper-dense cores in the training data, Suwon features a diverse mix of mid-rise residences, low-rise industry, and historical structures, providing a challenging out-of-distribution benchmark. All input features follow the same pre-processing pipeline as Sec.~2, while reference labels were sourced from Korea's national 3D mapping program. Notably, building height attributes are incomplete: only 30\% of footprints have valid height entries, highlighting the practical challenge of sparse supervision in new cities.

\begin{figure}[!t]
  \centering
  \includegraphics[width=\linewidth]{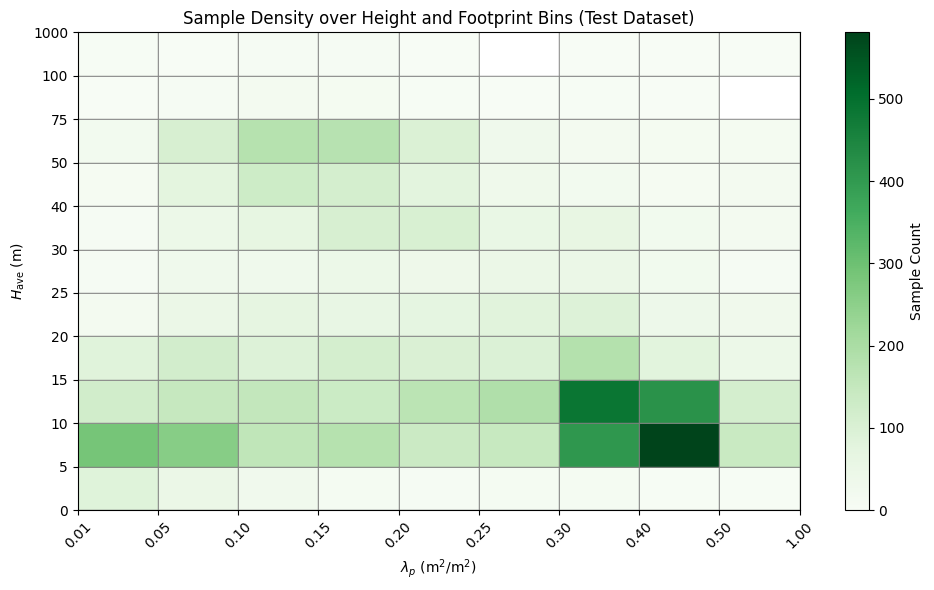}
  \caption{Joint distribution of building height and footprint in Suwon.}
  \label{fig:sample-distribution-suwon}
\end{figure}

According to a nationwide statistical analysis of 229 administrative regions in South Korea, the average building height is 10.07~m~\cite{LeeMin-Hyung2021}. To construct a representative evaluation subset, we analysed the joint distribution of building height (BH) and building footprint (BF) within Suwon. As shown in Fig.~\ref{fig:sample-distribution-suwon}, the majority of samples are concentrated within the BH range of 5--15~m and the BF range of 0.01--0.5. The BH range was chosen to avoid the influence of large deviations caused by extremely tall buildings with missing or inaccurate labels, while the BF range captures the densest portion of the distribution, ensuring statistical robustness. This subset reflects mid-rise, moderately dense urban structures and serves as a practical testbed for evaluating the model's generalisation capability.

\begin{figure}[!t]
  \centering
  \includegraphics[width=\linewidth]{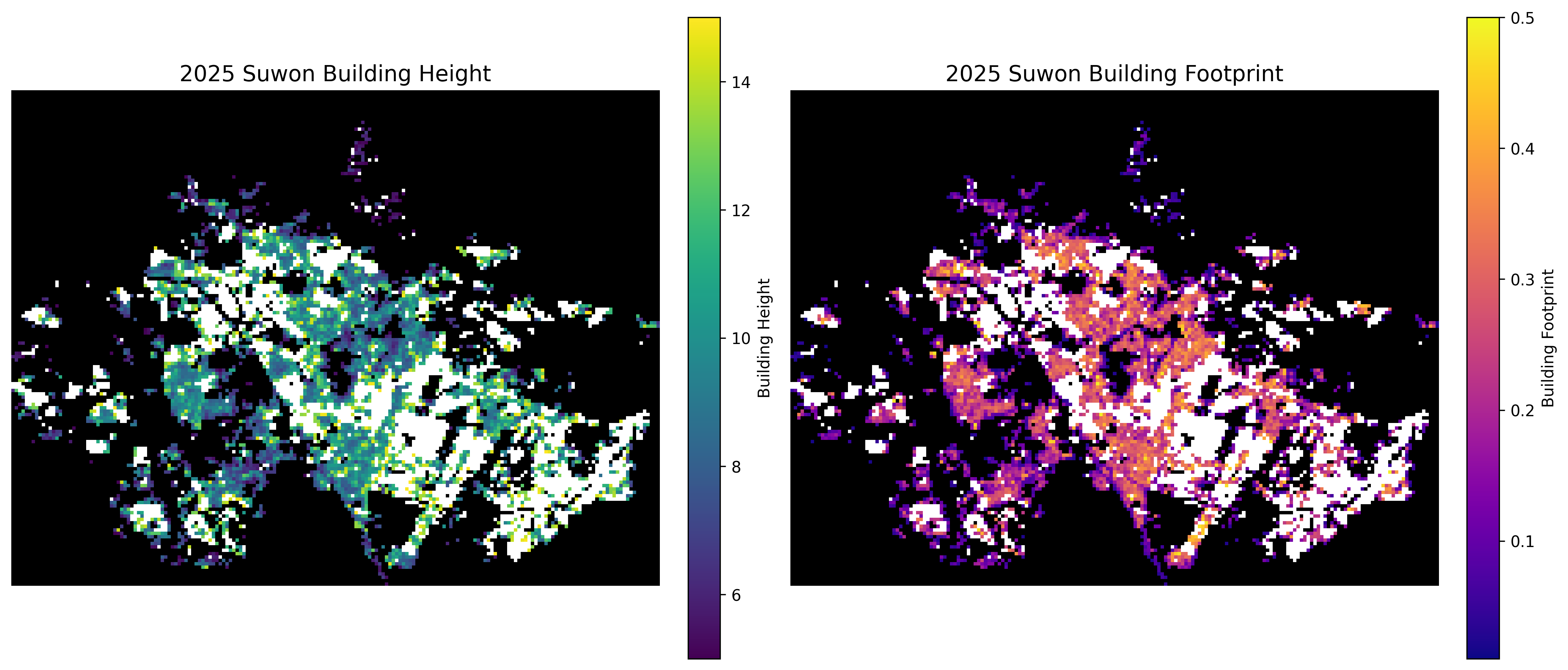}
  \caption{Spatial distribution of building height (left) and footprint (right) in Suwon for the evaluation subset ($5 \le H_{\mathrm{ave}} < 15$\,m and $0.01 \le \lambda_p < 0.5$). Colored pixels indicate selected grid cells.}
  \label{fig:suwon-spatial}
\end{figure}

Fig.~\ref{fig:suwon-spatial} illustrates the spatial distribution of building height (left) and footprint (right) in Suwon for the selected evaluation subset. The colored pixels denote grid cells that meet the selection criteria of $5 \le H_{\mathrm{ave}} < 15$\,m and $0.01 \le \lambda_p < 0.5$. This subset covers both central and peripheral regions of the city, ensuring a diverse morphological composition. 

\begin{table}[!t]
    \centering
    \caption{Model performance on the Suwon subset
             defined by $0.01 \le \lambda_{p} < 0.5$ and
             $5 \le H_{\mathrm{ave}} < 15$\,m.}
    \label{tab:suwon-subset}
    \begin{tabular}{lcc}
        \toprule
        Metric & Building Height & Building Footprint \\ \midrule
        RMSE   & 3.573\,m & 0.126 \\
        MAE    & 2.655\,m & 0.100 \\ \midrule
        \# Samples & \multicolumn{2}{c}{3\,710 \; (62.55\,\% of total)} \\
        \bottomrule
    \end{tabular}
\end{table}

We evaluated the best-performing GeoFormer model variant (5×5 contextual window) on this subset. As shown in Table~\ref{tab:suwon-subset}, the model achieved an RMSE of 3.573\,m for building height and 0.126 for building footprint, across 3,710 samples (62.55\% of the test set). These results validate the model's capacity to generalize across urban typologies in Suwon.

This geographically balanced sampling strategy enables a more rigorous evaluation of cross-domain generalization. By including both dense residential blocks and low-rise peripheral areas, it ensures that model performance is not inflated by spatially homogeneous regions. More importantly, the strong results on this subset demonstrate that the model maintains high accuracy even when applied to an \emph{unseen city} exhibiting distinct morphological and sensor characteristics compared to the training domains. This highlights the model's ability to generalize beyond distributional boundaries, validating its robustness under real-world domain shifts and spatial heterogeneity.

\subsection{Exploring Model Potential in an Earthquake-Affected City}

To further test GeoFormer's transferability, we apply the trained model---without any fine-tuning---to Marash (Kahramanmara\c{s}), one of the cities most heavily damaged by the 2023 Kahramanmaras earthquakes in southern Turkey. Paired Sentinel-1/2 and DEM inputs acquired within one month before and after the event were processed on a 100\,m grid under the same pipeline described in Section~II. This zero-shot deployment tests whether the learned building-parameter representations remain informative under extreme domain shift.

\begin{figure}[!t]
    \centering
    \includegraphics[width=\linewidth]{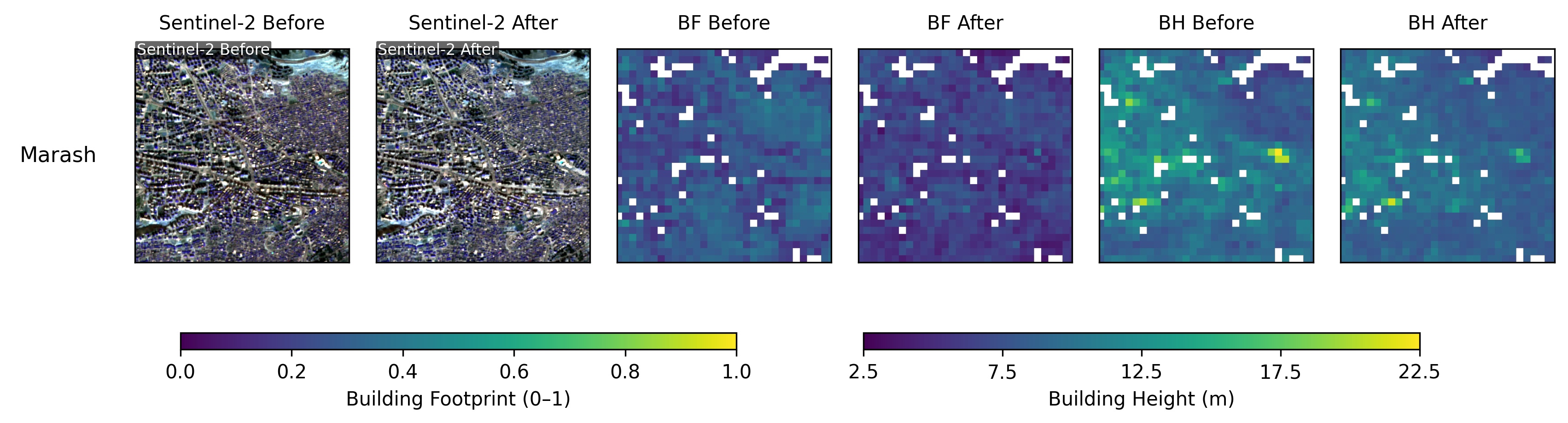}
    \caption{Pre- and post-earthquake Sentinel-2 imagery and predicted BF/BH for Marash}
    \label{fig:bf_bh_comparison}
\end{figure}

Fig.~\ref{fig:bf_bh_comparison} compares the predicted BF and BH maps for Marash before and after the earthquake. The post-event maps show both visibly reduced building coverage (BF) and lower predicted building heights (BH) in the city core, consistent with large-scale structural collapse documented by independent SAR-based damage assessments of the same earthquake sequence~\cite{ge2023sar_turkey,liu2024turkey_multisource}. This qualitative agreement---obtained without any disaster-specific training data or fine-tuning---demonstrates that GeoFormer's learned representations of urban morphology transfer to an extreme out-of-distribution scenario and remain sensitive to large-scale structural change.

We note that this case study is \emph{exploratory} and limited to qualitative visual comparison, as no building-level ground truth is available for the post-earthquake scene. Rigorous quantitative assessment would require coupling model predictions with building-level damage inventories and multi-temporal change detection, which we identify as a direction for future work.

\section{Conclusion}

This study presented GeoFormer, a lightweight Swin Transformer-based multi-task learning framework that jointly estimates building height (BH) and building footprint (BF) at 100\,m resolution using only freely available Sentinel-1 SAR, Sentinel-2 multispectral, and SRTM DEM data. Operating at 100\,m---the native resolution of major global building-parameter products---GeoFormer eliminates the need for proprietary VHR imagery, commercial satellite data, or ancillary vector layers, thereby ensuring full global reproducibility.

To ensure rigorous evaluation, we introduced GeoSplit, a geo-blocked spatial partitioning strategy that divides each city into radial sectors and allocates them to training, validation, and test sets with strict spatial independence. Unlike random sampling, GeoSplit prevents data leakage that would otherwise arise when context windows overlap between training and test patches, and preserves both central and peripheral urban morphologies in every subset. This design enables fair comparison across different context window sizes without conflating spatial proximity with model generalization.

Through systematic benchmarking against representative CNN architectures (ResNet-18, UNet, SENet) under identical training conditions across 54 morphologically diverse cities spanning four continents, we draw the following conclusions. First, GeoFormer achieves a BH RMSE of 3.19\,m with only 0.32\,M parameters---outperforming the best CNN baseline (UNet) by 7.5\,\% while requiring 35$\times$ fewer parameters than ResNet-18---demonstrating that windowed local attention is more effective than global convolution for scene-level building-parameter retrieval. Second, a 5$\times$5 context window (500\,m) provides the optimal trade-off between accuracy and efficiency; larger windows degrade performance through over-smoothing, and naively enlarging model capacity leads to overfitting rather than improved generalization. Third, ablation on input modality reveals that DEM is indispensable for height estimation ($+$15.0\,\% RMSE degradation upon removal), multispectral reflectance carries the dominant predictive signal ($+$37.9\,\% BH RMSE degradation upon removal), and the full SAR+Optical+DEM configuration remains optimal, confirming the complementary value of multi-source fusion.

Beyond in-sample evaluation, cross-dataset transfer to Suwon, Korea---a city entirely unseen during training---yielded a BH RMSE below 3.5\,m, confirming robust spatial generalization across markedly different urban morphologies. An exploratory zero-shot deployment over Kahramanmara\c{s} (Marash), a city heavily affected by the 2023 T\"{u}rkiye--Syria earthquake, further showed that GeoFormer's predicted footprint maps visually capture large-scale structural loss without any disaster-specific fine-tuning, pointing toward potential applications in rapid post-disaster building damage assessment.

Building on these results, future work will focus on extending the training set to additional cities and continents to produce a routinely updated global 100\,m BH/BF product, and on incorporating multi-temporal Sentinel sequences to enable monitoring of urban growth and post-disaster structural change over time.

\section*{Acknowledgment}
This research was supported by the 2023-MOIS36-004 (RS-2023-00248092) of the Technology Development Program on Disaster Restoration Capacity Building and Strengthening funded by the Ministry of Interior and Safety (MOIS, Korea).

\section*{Data Availability Statement}
The source code for the GeoFormer framework is publicly available on GitHub at
\url{https://github.com/geumjin99/geoformer}.
The building height and footprint rasters at 100\,m resolution are from the
SHAFTS dataset~\cite{shafts2023} (\url{https://zenodo.org/records/6370003}).

\section*{Declaration of Competing Interests}
The authors declare that they have no known competing financial interests or personal relationships that could have appeared to influence the work reported in this paper.

\bibliographystyle{IEEEtran}
\bibliography{cas-refs,new_refs}

\begin{thebibliography}{10}
\providecommand{\url}[1]{#1}
\csname url@samestyle\endcsname
\providecommand{\newblock}{\relax}
\providecommand{\bibinfo}[2]{#2}
\providecommand{\BIBentrySTDinterwordspacing}{\spaceskip=0pt\relax}
\providecommand{\BIBentryALTinterwordstretchfactor}{4}
\providecommand{\BIBentryALTinterwordspacing}{\spaceskip=\fontdimen2\font plus
\BIBentryALTinterwordstretchfactor\fontdimen3\font minus
  \fontdimen4\font\relax}
\providecommand{\BIBforeignlanguage}[2]{{%
\expandafter\ifx\csname l@#1\endcsname\relax
\typeout{** WARNING: IEEEtran.bst: No hyphenation pattern has been}%
\typeout{** loaded for the language `#1'. Using the pattern for}%
\typeout{** the default language instead.}%
\else
\language=\csname l@#1\endcsname
\fi
#2}}
\providecommand{\BIBdecl}{\relax}
\BIBdecl

\bibitem{fang2016changing}
C.~Fang, G.~Li, and S.~Wang, ``Changing and differentiated urban landscape in
  {{China}}: {{Spatiotemporal}} patterns and driving forces,'' \emph{Environ.
  Sci. Technol.}, vol.~50, no.~5, pp. 2217--2227, 2016.

\bibitem{frolking2013global}
S.~Frolking, T.~Milliman, K.~C. Seto, and M.~A. Friedl, ``A global fingerprint
  of macro-scale changes in urban structure from 1999 to 2009,'' \emph{Environ.
  Res. Lett.}, vol.~8, no.~2, p. 024004, 2013.

\bibitem{frolking2024global}
S.~Frolking, R.~Mahtta, T.~Milliman, T.~Esch, and K.~C. Seto, ``Global urban
  structural growth shows a profound shift from spreading out to building up,''
  \emph{Nat. Cities}, vol.~1, no.~9, pp. 555--566, 2024.

\bibitem{xi2021impacts}
C.~Xi, C.~Ren, J.~Wang, Z.~Feng, and S.-J. Cao, ``Impacts of urban-scale
  building height diversity on urban climates: {{A}} case study of {{Nanjing}},
  {{China}},'' \emph{Energy Build.}, vol. 251, p. 111350, 2021.

\bibitem{perini2014effects}
K.~Perini and A.~Magliocco, ``Effects of vegetation, urban density, building
  height, and atmospheric conditions on local temperatures and thermal
  comfort,'' \emph{Urban For. Urban Green.}, vol.~13, no.~3, pp. 495--506,
  2014.

\bibitem{huang2020estimates}
X.~Huang and C.~Wang, ``Estimates of exposure to the 100-year floods in the
  conterminous {{United States}} using national building footprints,''
  \emph{Int. J. Disaster Risk Reduct.}, vol.~50, p. 101731, 2020.

\bibitem{tian2025fire}
Y.~Tian, M.~Lu, Z.~Xu, and J.~Ren, ``A fire following earthquake spread model
  considering building height and its application to real-world events,''
  \emph{Int. J. Disaster Risk Reduct.}, p. 105261, 2025.

\bibitem{geofabrik_osm_2018}
{Geofabrik}, ``{{OpenStreetMap}} download statistics,'' 2018.

\bibitem{li2020developing}
X.~Li, Y.~Zhou, P.~Gong, K.~C. Seto, and N.~Clinton, ``Developing a method to
  estimate building height from {{Sentinel-1}} data,'' \emph{Remote Sens.
  Environ.}, vol. 240, p. 111705, 2020.

\bibitem{cai2023deep}
B.~Cai, Z.~Shao, X.~Huang, X.~Zhou, and S.~Fang, ``Deep learning-based building
  height mapping using {{Sentinel-1}} and {{Sentinel-2}} data,'' \emph{Int. J.
  Appl. Earth Obs. Geoinformation}, vol. 122, p. 103399, 2023.

\bibitem{frantz2021national}
D.~Frantz, F.~Schug, A.~Okujeni, C.~Navacchi, W.~Wagner, S.~{van der Linden},
  and P.~Hostert, ``National-scale mapping of building height using
  {{Sentinel-1}} and {{Sentinel-2}} time series,'' \emph{Remote Sens.
  Environ.}, vol. 252, p. 112128, 2021.

\bibitem{dabrock2024leveraging}
K.~Dabrock, N.~Pflugradt, J.~M. Weinand, and D.~Stolten, ``Leveraging machine
  learning to generate a unified and complete building height dataset for
  {{Germany}},'' \emph{Energy AI}, vol.~17, p. 100408, 2024.

\bibitem{buyukdemircioglu2022deep}
M.~Buyukdemircioglu, R.~Can, S.~Kocaman, and M.~Kada, ``Deep learning based
  building footprint extraction from very high resolution true orthophotos and
  {{nDSM}},'' \emph{ISPRS Ann. Photogramm. Remote Sens. Spat. Inf. Sci.},
  vol.~2, pp. 211--218, 2022.

\bibitem{li2021deep}
Z.~Li, Q.~Xin, Y.~Sun, and M.~Cao, ``A deep learning-based framework for
  automated extraction of building footprint polygons from very high-resolution
  aerial imagery,'' \emph{Remote Sens.}, vol.~13, no.~18, p. 3630, 2021.

\bibitem{shafts2023}
R.~Li, T.~Sun, F.~Tian, and G.-H. Ni, ``{{SHAFTS}} (v2022.3): A
  deep-learning-based {{Python}} package for simultaneous extraction of
  building height and footprint from {{Sentinel}} imagery,'' \emph{Geosci.
  Model Dev.}, vol.~16, no.~2, pp. 751--778, 2023.

\bibitem{rastogi2022automatic}
K.~Rastogi, P.~Bodani, and S.~A. Sharma, ``Automatic building footprint
  extraction from very high-resolution imagery using deep learning
  techniques,'' \emph{Geocarto Int.}, vol.~37, no.~5, pp. 1501--1513, 2022.

\bibitem{park2019creating}
Y.~Park and J.-M. Guldmann, ``Creating {{3D}} city models with building
  footprints and {{LIDAR}} point cloud classification: {{A}} machine learning
  approach,'' \emph{Comput. Environ. Urban Syst.}, vol.~75, pp. 76--89, 2019.

\bibitem{sun2019large}
Y.~Sun, Y.~Hua, L.~Mou, and {\relax XX}.~Zhu, ``Large-scale building height
  estimation from single {{VHR SAR}} image using fully convolutional network
  and {{GIS}} building footprints. 2019 {{Joint Urban Remote Sensing Event}},
  {{JURSE}} 2019,'' 2019.

\bibitem{cai2024automated}
P.~Cai, J.~Guo, R.~Li, Z.~Xiao, H.~Fu, T.~Guo, X.~Zhang, Y.~Li, and X.~Song,
  ``Automated building height estimation using ice, cloud, and land elevation
  satellite 2 light detection and ranging data and building footprints,''
  \emph{Remote Sens.}, vol.~16, no.~2, p. 263, 2024.

\bibitem{wu2023first}
W.-B. Wu, J.~Ma, E.~Banzhaf, M.~E. Meadows, Z.-W. Yu, F.-X. Guo, D.~Sengupta,
  X.-X. Cai, and B.~Zhao, ``A first {{Chinese}} building height estimate at 10
  m resolution ({{CNBH-10}} m) using multi-source earth observations and
  machine learning,'' \emph{Remote Sens. Environ.}, vol. 291, p. 113578, 2023.

\bibitem{chen2024refining}
Y.~Chen, W.~Sun, L.~Yang, X.~Yang, X.~Zhou, X.~Li, S.~Li, and G.~Tang,
  ``Refining urban morphology: {{An}} explainable machine learning method for
  estimating footprint-level building height,'' \emph{Sustain. Cities Soc.},
  vol. 112, p. 105635, 2024.

\bibitem{chen2025structure}
Y.~Chen, J.~Zhou, C.~Xu, Q.~Ma, X.~Zhang, Y.~Zhou, and Y.~Ge, ``Structure-aware
  deep learning network for building height estimation,'' \emph{Int. J. Appl.
  Earth Obs. Geoinformation}, p. 104443, 2025.

\bibitem{che20243d}
Y.~Che, X.~Li, X.~Liu, Y.~Wang, W.~Liao, X.~Zheng, X.~Zhang, X.~Xu, Q.~Shi,
  J.~Zhu \emph{et~al.}, ``{{3D-GloBFP}}: {{The}} first global three-dimensional
  building footprint dataset,'' \emph{Earth Syst. Sci. Data Discuss.}, vol.
  2024, pp. 1--28, 2024.

\bibitem{wang2024mfbhnet}
S.~Wang, B.~Cai, D.~Hou, Q.~Ding, J.~Wang, and Z.~Shao, ``Mf-bhnet: A hybrid
  multimodal fusion network for building height estimation using sentinel-1 and
  sentinel-2 imagery,'' \emph{IEEE Transactions on Geoscience and Remote
  Sensing}, vol.~62, pp. 1--19, 2024.

\bibitem{zheng2025nectnet}
Y.~Zheng \emph{et~al.}, ``Estimating individual building heights by integrating
  spaceborne {LiDAR} and multisource remote sensing data: A {CNN}--transformer
  model and a semi-supervised sample augmentation approach,'' \emph{IEEE
  Transactions on Geoscience and Remote Sensing}, vol.~63, 2025.

\bibitem{mostafavi2024utglobus}
H.~G. Kamath, M.~Singh, N.~Malviya, A.~Martilli, L.~He, D.~Aliaga, C.~He,
  F.~Chen, L.~A. Magruder, Z.-L. Yang \emph{et~al.}, ``Global building heights
  for urban studies (ut-globus) for city-and street-scale urban simulations:
  Development and first applications,'' \emph{Scientific Data}, vol.~11, no.~1,
  p. 886, 2024.

\bibitem{zhu2025globalbuildingatlas}
X.~Zhu, S.~Chen, F.~Zhang, Y.~Shi, and Y.~Wang, ``{GlobalBuildingAtlas}: An
  open global and complete dataset of building polygons, heights and {LoD1}
  {3D} models,'' \emph{Earth System Science Data}, vol.~17, pp. 6647--6670,
  2025.

\bibitem{fisher1997pixel}
P.~Fisher, ``The pixel: A snare and a delusion,'' \emph{Int. J. Remote Sens.},
  vol.~18, no.~3, pp. 679--685, 1997.

\bibitem{weng2012remote}
Q.~Weng, ``Remote sensing of impervious surfaces in the urban areas:
  Requirements, methods, and trends,'' \emph{Remote Sens. Environ.}, vol. 117,
  pp. 34--49, 2012.

\bibitem{brunner2010earthquake}
D.~Brunner, G.~Lemoine, and L.~Bruzzone, ``Earthquake damage assessment of
  buildings using {VHR} optical and {SAR} imagery,'' \emph{IEEE Trans. Geosci.
  Remote Sens.}, vol.~48, no.~5, pp. 2403--2420, 2010.

\bibitem{schug2022sub}
F.~Schug, D.~Frantz, A.~Okujeni, and P.~Hostert, ``Sub-pixel building area
  mapping based on synthetic training data and regression-based unmixing using
  {{Sentinel-1}} and-2 data,'' \emph{Remote Sens. Lett.}, vol.~13, no.~8, pp.
  822--832, 2022.

\bibitem{radoux2016sentinel}
J.~Radoux, G.~Chom{\'e}, D.~C. Jacques, F.~Waldner, N.~Bellemans, N.~Matton,
  C.~Lamarche, R.~{d'Andrimont}, and P.~Defourny, ``Sentinel-2's potential for
  sub-pixel landscape feature detection,'' \emph{Remote Sens.}, vol.~8, no.~6,
  p. 488, 2016.

\bibitem{stewart2012local}
I.~D. Stewart and T.~R. Oke, ``Local climate zones for urban temperature
  studies,'' \emph{Bull. Am. Meteorol. Soc.}, vol.~93, no.~12, pp. 1879--1900,
  2012.

\bibitem{demuzere2022global}
M.~Demuzere, J.~Kittner, A.~Martilli, G.~Mills, C.~Moede, I.~D. Stewart,
  J.~Van~Vliet, and B.~Bechtel, ``A global map of local climate zones to
  support earth system modelling and urban scale environmental science,''
  \emph{Earth System Science Data Discussions}, vol. 2022, pp. 1--57, 2022.

\bibitem{ching2018wudapt}
J.~Ching, G.~Mills, B.~Bechtel, L.~See, J.~Feddema, X.~Wang, C.~Ren,
  O.~Brousse, A.~Martilli, M.~Neophytou \emph{et~al.}, ``{WUDAPT}: An urban
  weather, climate, and environmental modeling infrastructure for the
  anthropocene,'' \emph{Bull. Am. Meteorol. Soc.}, vol.~99, no.~9, pp.
  1907--1924, 2018.

\bibitem{bechtel2015mapping}
B.~Bechtel, P.~J. Alexander, J.~B{\"o}hner, J.~Ching, O.~Conrad, J.~Feddema,
  G.~Mills, L.~See, and I.~Stewart, ``Mapping local climate zones for a
  worldwide database of the form and function of cities,'' \emph{ISPRS Int. J.
  Geo-Inf.}, vol.~4, no.~1, pp. 199--219, 2015.

\bibitem{martilli2002urban}
A.~Martilli, A.~Clappier, and M.~W. Rotach, ``An urban surface exchange
  parameterisation for mesoscale models,'' \emph{Bound.-Layer Meteorol.}, vol.
  104, pp. 261--304, 2002.

\bibitem{tatem2017worldpop}
A.~J. Tatem, ``{WorldPop}, open data for spatial demography,'' \emph{Sci.
  Data}, vol.~4, p. 170004, 2017.

\bibitem{schiavina2023ghspop}
M.~Schiavina, S.~Freire, A.~Carioli, and K.~MacManus, ``{GHS-POP R2023A} --
  {GHS} population grid multitemporal (1975--2030),'' European Commission,
  Joint Research Centre (JRC), 2023, available at 100\,m resolution.

\bibitem{reinhart2016urban}
C.~F. Reinhart and C.~{Cerezo Davila}, ``Urban building energy modeling -- a
  review of a nascent field,'' \emph{Build. Environ.}, vol.~97, pp. 196--202,
  2016.

\bibitem{musiaka2021}
{\L}.~Musiaka and M.~Nalej, ``Application of {{GIS}} tools in the measurement
  analysis of urban spatial layouts using the square grid method,'' \emph{ISPRS
  Int. J. Geo-Inf.}, vol.~10, no.~8, p. 558, 2021.

\bibitem{liu2021swin}
Z.~Liu, Y.~Lin, Y.~Cao, H.~Hu, Y.~Wei, Z.~Zhang, S.~Lin, and B.~Guo, ``Swin
  transformer: {{Hierarchical}} vision transformer using shifted windows,'' in
  \emph{Proc. {{IEEECVF Int}}. {{Conf}}. {{Comput}}. {{Vis}}.}, 2021, pp.
  10\,012--10\,022.

\bibitem{kendall2018multi}
A.~Kendall, Y.~Gal, and R.~Cipolla, ``Multi-task learning using uncertainty to
  weigh losses for scene geometry and semantics,'' in \emph{Proc. {{IEEE
  Conf}}. {{Comput}}. {{Vis}}. {{Pattern Recognit}}.}, 2018, pp. 7482--7491.

\bibitem{huber1964robust}
P.~J. Huber, ``Robust estimation of a location parameter,'' in
  \emph{Breakthroughs in Statistics: Methodology and Distribution}.\hskip 1em
  plus 0.5em minus 0.4em\relax Springer, 1992, pp. 492--518.

\bibitem{loshchilov2017decoupled}
I.~Loshchilov and F.~Hutter, ``Decoupled weight decay regularization,''
  \emph{ArXiv Prepr. ArXiv171105101}, 2017.

\bibitem{loshchilov2016sgdr}
------, ``Sgdr: {{Stochastic}} gradient descent with warm restarts,''
  \emph{ArXiv Prepr. ArXiv160803983}, 2016.

\bibitem{he2016deep}
K.~He, X.~Zhang, S.~Ren, and J.~Sun, ``Deep residual learning for image
  recognition,'' in \emph{Proc. {{IEEE Conf}}. {{Comput}}. {{Vis}}. {{Pattern
  Recognit}}.}, 2016, pp. 770--778.

\bibitem{ronneberger2015u}
O.~Ronneberger, P.~Fischer, and T.~Brox, ``{U-Net}: Convolutional networks for
  biomedical image segmentation,'' in \emph{Medical Image Computing and
  Computer-Assisted Intervention (MICCAI)}.\hskip 1em plus 0.5em minus
  0.4em\relax Springer, 2015, pp. 234--241.

\bibitem{hu2018squeeze}
J.~Hu, L.~Shen, and G.~Sun, ``Squeeze-and-excitation networks,'' in \emph{Proc.
  IEEE/CVF Conf. Computer Vision and Pattern Recognition (CVPR)}, 2018, pp.
  7132--7141.

\bibitem{LeeMin-Hyung2021}
M.-H. Lee, W.-S. Seo, C.-Y. Park, and C.-H. Choi, ``Improvement of surface
  roughness classification criteria reflecting the height and density of
  building by region,'' \emph{J. Korean Inst. Archit. Sustain. Environ. Build.
  Syst. KIAEBS}, vol.~15, no.~5, pp. 513--524, 2021.

\bibitem{ge2023sar_turkey}
X.~Wang, G.~Feng, L.~He, Q.~An, Z.~Xiong, H.~Lu, W.~Wang, N.~Li, Y.~Zhao,
  Y.~Wang, and Y.~Wang, ``Evaluating urban building damage of 2023
  {Kahramanmaras}, {Turkey} earthquake sequence using {SAR} change detection,''
  \emph{Sensors}, vol.~23, no.~14, p. 6342, 2023.

\bibitem{liu2024turkey_multisource}
X.~Yu, X.~Hu, Y.~Song \emph{et~al.}, ``Intelligent assessment of building
  damage of 2023 {Turkey}--{Syria} earthquake by multiple remote sensing
  approaches,'' \emph{npj Nat. Hazards}, vol.~1, p.~3, 2024.

\end{thebibliography}


\end{document}